\documentclass[lettersize,journal]{IEEEtran}
\usepackage{amsmath,amsfonts}
\usepackage{amsthm}
\usepackage{amssymb}
\usepackage{calligra}
\usepackage{algorithmic}
\usepackage{array}
\usepackage{bbm}
\usepackage[caption=false,font=normalsize,labelfont=sf,textfont=sf]{subfig}
\usepackage{textcomp}
\usepackage{stfloats}
\usepackage{url}
\usepackage{verbatim}
\usepackage{graphicx}
\def\BibTeX{{\rm B\kern-.05em{\sc i\kern-.025em b}\kern-.08em
    T\kern-.1667em\lower.7ex\hbox{E}\kern-.125emX}}
\usepackage{balance}
\usepackage{makecell}
\usepackage{multirow}
\usepackage{lipsum}
\usepackage{mathrsfs}

\newtheorem{theorem}{Theorem}

\newtheorem{proof outline}{Proof Outline}[theorem]
\newtheorem*{remarks}{Remarks}

\begin{document}
\title{Representation and De-interleaving of Mixtures of Hidden Markov Processes}
\author{Jiadi Bao,~\IEEEmembership{Student Member,~IEEE,}
Mengtao Zhu,~\IEEEmembership{Member,~IEEE,}\\
Yunjie Li,~\IEEEmembership{Senior Member,~IEEE,}
and Shafei Wang
\thanks{This work was supported by the National Natural Science Foundation (NSFC) of China under grant no. 62301031, Young Elite Scientists Sponsorship Program by BAST under grant BYESS2023415. \textit{(corresponding author: Yunjie Li)}.}
\thanks{Jiadi Bao and Yunjie Li are with the School of Information and Electronics, Beijing Institute of Technology, Beijing, 100081, China (E-mail: baojiadi@bit.edu.cn; liyunjie@bit.edu.cn).}
\thanks{Mengtao Zhu is with the School of Cyberspace Science and Technology, Beijing Institute of Technology, Beijing, 100081, China (E-mail: zhumengtao@bit.edu.cn). }
\thanks{Shafei Wang is with the School of Information and Electronics, Beijing Institute of Technology, Beijing, 100081, China, and also with the Laboratory of Electromagnetic Space Cognition and Intelligent Control, Beijing, 100191, China.}}

\markboth{Journal of \LaTeX\ Class Files,~Vol.~18, No.~9, April~2024}%
{How to Use the IEEEtran \LaTeX \ Templates}

\maketitle

\begin{abstract}
De-interleaving of the mixtures of Hidden Markov Processes (HMPs) generally depends on its representation model. Existing representation models consider Markov chain mixtures rather than hidden Markov, resulting in the lack of robustness to non-ideal situations such as observation noise or missing observations. Besides, de-interleaving methods utilize a search-based strategy, which is time-consuming. To address these issues, this paper proposes a novel representation model and corresponding de-interleaving methods for the mixtures of HMPs. At first, a generative model for representing the mixtures of HMPs is designed. Subsequently, the de-interleaving process is formulated as a posterior inference for the generative model. Secondly, an exact inference method is developed to maximize the likelihood of the complete data, and two approximate inference methods are developed to maximize the evidence lower bound by creating tractable structures. Then, a theoretical error probability lower bound is derived using the likelihood ratio test, and the algorithms are shown to get reasonably close to the bound. Finally, simulation results demonstrate that the proposed methods are highly effective and robust for non-ideal situations, outperforming baseline methods on simulated and real-life data.
\end{abstract}


\section{Introduction}
\begin{figure}[!t]
  \centering
  \includegraphics[width=3.6in]{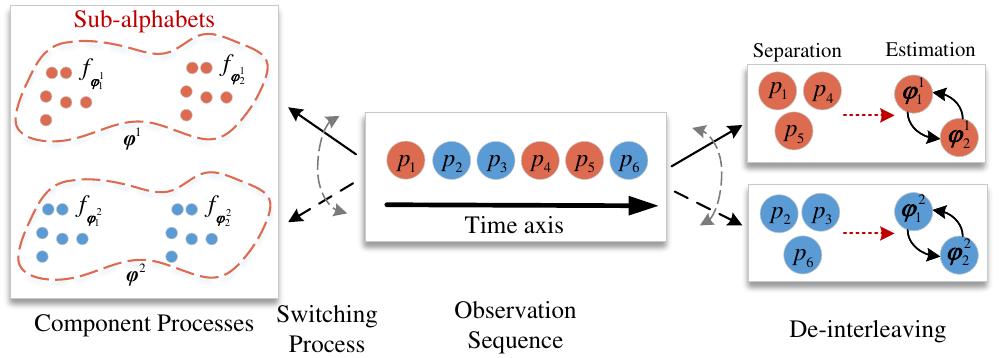}
  \caption{Schematic setup for two independent sources.}
  \label{IHMP}
\end{figure}

\IEEEPARstart{H}{idden} Markov process (HMP) \cite{ephraim2002hidden} is a discrete-time finite-state homogeneous Markov chain observed through a discrete-time memory-less invariant channel. HMPs are commonly modeled as hidden Markov models (HMMs) and HMM has been wildly used in modeling and analysis time-series, such as data mining \cite{Qiao2015}, radar signal recognition \cite{wang2008signal, Bao2023}, target recognition \cite{Du2011bayesian}, human activity recognition \cite{Gu2010unsupervised}, etc. However, in some asynchronous \cite{wang2018signal} or anonymized \cite{krishnamurthy2023adaptive} systems, the observation of multiple independent time-series is interleaved (mixed). The interleaved observation brings great challenges for time-series pattern recognition and parameter estimation. To overcome the challenges, in this paper, we consider the following problem:

There are $M$ independent sources emitting signals. Each source is assumed to have finite-memory (Markov). The observed signal is contaminated by noise, so the observed signal can be considered generated from a random distribution. In such circumstances, a signal emitted by a source is an HMP. This HMP emitted by a source is also referred to as \textit{component process}, and is generated by a Markov chain called \textit{component chain}. Let $\mathcal{A}=\boldsymbol{\varphi}^1\cup\boldsymbol{\varphi}^2\cup...\boldsymbol{\varphi}^M$ be the finite-alphabet, and $\boldsymbol{\Pi}=\{\boldsymbol{\varphi}^1, ..., \boldsymbol{\varphi}^M\}$ be the partition of the alphabet. Let $\boldsymbol{\varphi}^m=\{\boldsymbol{\varphi}_1^m,...,\boldsymbol{\varphi}_k^m,...,\boldsymbol{\varphi}_{K^m}^m\},m\in[1,M]$ be the finite, non-empty sub-alphabet of $\mathcal{A}$ emitted by the $m$th component chain, where $K^m$ is the symbol number of the $m$th sub-alphabet, also corresponds to the hidden states number of the $m$th component chain. From a generative perspective of the interleaved observation sequence with length $T$: $\boldsymbol{p}=\{\boldsymbol{p}_t\}_{t=1}^T$. The component processes are generated according to the symbol $\boldsymbol{\varphi}_k^m$. The symbols are considered random variables with the probability density function $f_{\boldsymbol{\varphi}_m^k}(\cdot)$. Additionally, the component processes are interleaved by an additional random process called \textit{switching process}. Further, the switching process is modeled by a Markov chain called \textit{switching chain}. In conclusion, the objective of the de-interleaving mixtures of HMPs is two-fold: firstly, to separate the component processes generated by the different sources; secondly, to accurately estimate the parameters of the component chains and symbols. The schematic diagram of this process is depicted in Fig.~\ref{IHMP}. 

The first step of de-interleaving HMP mixtures focuses on formulating the underlying generative model of the interleaved observation sequence. In~\cite{Batu2004}, the authors first described the interleaved Markov chains and further summarized by Seroussi \cite{Seroussi2009,Seroussi2011,Seroussi2012} as the Interleaved Markov Process (IMP), in which the sub-alphabets are assumed to be disjoint. In addition, a variant of IMP was also investigated \cite{Gillblad2009,Minot2014}, where the sub-alphabets are considered non-disjoint. The above studies have limited the component processes to be Markov rather than hidden Markov. In a subsequent development \cite{Landwehr2008}, the authors extend the interleaved scheme to hidden Markov processes. However, the observations are assumed to be discrete which makes the model rather limited facing the observation noise and missing observations. Thus, a new generative model with better representation capabilities needs to be designed.


The second step of de-interleaving HMP mixtures is to infer the hidden variables of the generative model and estimate the model parameters. Major investigations focus on exhaustive search or heuristic optimization methods to perform inference \cite{Seroussi2012, Gabriel2020, Pinsolle2023, pinsolle2024deinterleaving}. Generally, there are two main drawbacks of the above methods. Firstly, only the order of each component chain $\boldsymbol{k}$ and the partition $\boldsymbol{\Pi}$ are considered as objectives, but the hidden state assignments are ignored in the existing researches. Hidden state assignments are crucial for source behavior recognition \cite{Fox2011}, mode change discovery \cite{Bao2023}, and many other applications. Secondly, these methods are time-consuming since finding alphabet partition is a combinatorial optimization problem. Alternatively, an Expectation Maximization (EM) algorithm is described in \cite{Minot2014} to learn the model parameter and utilize naive Viterbi to perform the hidden state inference under disjoint and non-disjoint sub-alphabets. However, the computational time of each iteration grows exponentially as the alphabet size $|\mathcal{A}|$ grows. Further, the above investigations do not consider non-ideal conditions such as observation noise and missing observations. Recently, with the development of deep learning, the supervised method \cite{Zhu2022} was designed to de-interleave Markov processes under non-ideal conditions. However, note that supervised methods require prior labeled data for network training. Thus, three problems remain to be solved for de-interleaving HMPs: 1) An underlying generative model needs to be designed to model the HMP mixtures, and the model is required to meet practical applications. 2) The unsupervised de-interleaving algorithm has to be equipped with efficiency. 3) The designed algorithm demands robust to non-ideal conditions such as observation noise and missing observations when the component chain emits either disjoint or non-disjoint sub-alphabets.

Taking the above problems into consideration, this paper proposes an efficient and unsupervised method to de-interleave the mixtures of HMPs. Firstly, the Interleaved Hidden Markov Process (IHMP) is introduced to describe the HMP mixtures, and a generative model is designed for IHMP modeling. Secondly,  de-interleave HMP mixtures are treated as posterior inference for the generative model. Specifically, given a sequence of observations, find the most likely configuration of hidden variables to have generated the observed data. Then, an exact inference method based on the EM algorithm is proposed, while the exact inference is shown to be NP-hard due to its combinatorial nature. Alternatively, the approximate inference is utilized. Variational Inference (VI) \cite{Zhang2018, Blei2017} is an efficient way to approximate the intractable posterior. The primary mechanism of VI is to minimize the Kullback-Leibler (KL) \cite{Kullback1951} divergence between the variational distribution $Q$ and the true posterior distribution $P$. The expensive iterative inference schemes in another approximate inference method (Markov chain Monte Carlo (MCMC) \cite{Jones2022markov}) are avoided. Finally, an error probability lower bound is derived based on the likelihood ratio test to determine how close is our proposed algorithms to the optimum. Simulations and two applications verified the effectiveness of the proposed method. The main contributions of our work are summarized as follows:
\begin{enumerate}
  \item A generative model is proposed to model the IHMP. The proposed model is more suitable for modeling the time-series contaminated by the noise. In the context of the IHMP, the search space of the alphabet partition is much smaller than the search space presented in the previous IMP paper \cite{Seroussi2012}, i.e., when there are 10 symbols, the search space of IMP and IHMP are 115975 and 42, respectively.
  \item Three inference algorithms with different levels of variable coupling are proposed. Three methods are the exact (EM) algorithm \cite{Dempster1977}, variational inference based on mean-field approximation \cite{Lawrence1996}, and variational inference based on structured approximation \cite{Ghahramani1997}. The update function of the proposed algorithms is explicitly derived.
  \item A theoretical error probability lower bound on de-interleaving two binary-state HMMs is derived. The simulations show that the proposed methods are reasonably close to the bound.
\end{enumerate}

The rest of this paper is organized as follows. Section II presents a review of the hidden Markov model with Gaussian emission and the interleaved Markov process. Section III illustrates the probabilistic formulation of the generative model for the IHMP, and the combinatorial problem between IHMP and IMP is discussed. Section IV proposed three methods of de-interleaving HMPs with different levels of variable couplings, and the derivation of this section is presented in Appendix A to C. Section V provides the error probability of separating two binary-state HMMs, the derivation of this section is presented in Appendix D. Simulated results demonstrating the effectiveness are provided in section VI. Section VII shows the application results on radar data and human motion data. Section VIII concludes the paper.

\section{Preliminary}

In this section, the HMM with Gaussian emissions \cite{Kontorovich2013} and the IMP \cite{Seroussi2012} are reviewed.

\subsection{Hidden Markov Model with Gaussian Emissions}
\begin{figure*}[!t]
\centering
\includegraphics[width=6.8in]{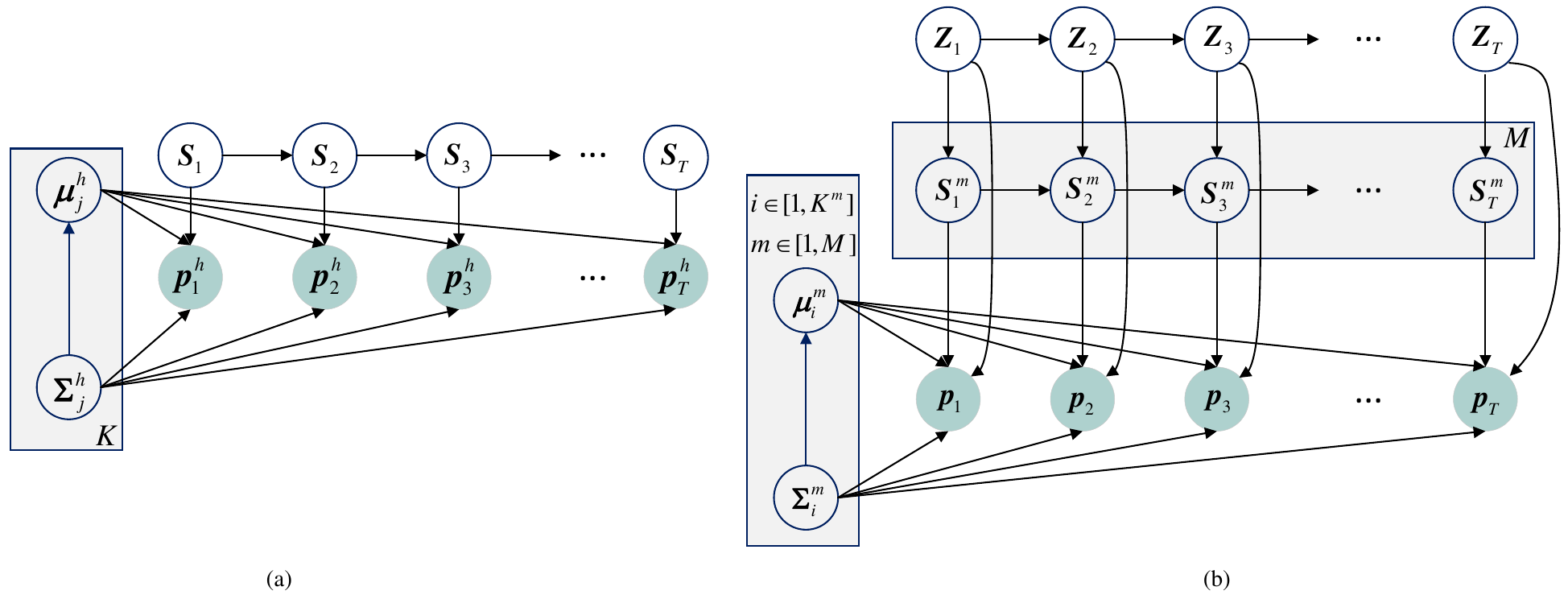}
\caption{(a) The HMM with multi-variate Gaussian emission. (b) The generative model of IHMP.}
\label{graphical model}
\end{figure*}

The HMM with Gaussian emissions is characterized by a three-tuple:
\begin{equation}
    \boldsymbol{\lambda}^h = (\boldsymbol{\pi}^h,\boldsymbol{A}^h,\boldsymbol{\varphi}^h)
\end{equation}
where the superscript $h$ is short for HMM, $\boldsymbol{\pi}^h = \{\pi_i^h\}_{i=1}^K$ is the prior distribution with $K$ elements, $\boldsymbol{A}^h = \{\boldsymbol{A}_j^h\}_{j=1}^K$,$\boldsymbol{A}_j^h=\{A_{j,i}^h\}_{i=1}^K$ is the state transition matrix with size $K\times K$. There are $K$ hidden states and each hidden state corresponds to a random variable following the Gaussian distribution characterized by  $\boldsymbol{\varphi}^h_i\sim \mathcal{N}(\boldsymbol{\mu}_i^h,\boldsymbol{\Sigma}_i^h)$, where $\mathcal{N}$ represents the Gaussian distribution,  $\boldsymbol{\mu}_i^h$ is the mean of the Gaussian distribution and $\boldsymbol{\Sigma}_i^h$ is the covariance matrix of the Gaussian distribution. These Gaussian random variables construct the model set $\boldsymbol{\varphi}^h=\{\boldsymbol{\varphi}^h_i\}_{i=1}^K$. The probability density function of a Gaussian random variable is defined as:
\begin{equation}
\label{GaussianDistribution}
\begin{aligned}
    &f_{\boldsymbol{\varphi}^h_i}(\boldsymbol{p}_t^h)=\\
    &\frac{1}{\boldsymbol{C}^h} \exp\left(-\frac{1}{2}\left(\boldsymbol{p}_t^h-\boldsymbol{\mu}_i^h\right)^\top\left(\boldsymbol{\Sigma}_i^h\right)^{-1}\left(\boldsymbol{p}_t^h-\boldsymbol{\mu}_i^h\right)\right)
\end{aligned}
\end{equation}
where $\boldsymbol{C}^h$ is the normalization term, $\boldsymbol{v}^\top$ represents the transpose of the vector $\boldsymbol{v}$,  and $\boldsymbol{p}_t^h$ is the observed variable of HMM at time instant $t$. The probabilistic graphical model of the HMM with Gaussian emissions is shown in Fig. \ref{graphical model}(a). 

\subsection{Interleaved Markov Process}
The interleaved Markov process is described in paper \cite{Seroussi2012}. The interleaved symbol sequence with length $T$ is denoted as $\boldsymbol{p}=\{\boldsymbol{p}_t\}_{t=1}^T$. Firstly, let $\boldsymbol{A_\Pi}(\boldsymbol{p})\in \boldsymbol{\Pi}$ denotes the corresponding sequence of the sub-alphabets, i.e., $\boldsymbol{A_\Pi}(\boldsymbol{p})_j=\boldsymbol{\varphi}^i$, where $i$ is the unique index such that $\boldsymbol{p}_j \in \boldsymbol{\varphi}^i\in \boldsymbol{\Pi}$,$ 1\leq j\leq T$, $\boldsymbol{\varphi}^i$ is the sub-alphabet and $\boldsymbol{\Pi}$ is the partition. The $\boldsymbol{A}_{\boldsymbol{\Pi}}(\boldsymbol{p})\in \boldsymbol{\Pi}$ can also be called a switching sequence corresponding to the observed symbol sequence. Secondly, Let $\boldsymbol{p}[\boldsymbol{\varphi}^i]$ denote the symbols sequences that all symbols in sub-alphabet $\boldsymbol{\varphi}^i$. The IMP is defined as follows: Given $\boldsymbol{p} =\{\boldsymbol{p}_t\}_{t=1}^T$, and assuming $\boldsymbol{p}_t\in \boldsymbol{\varphi}^i$, the likelihood is defined:
\begin{equation}
\label{IMP}
  P\left(\boldsymbol{p}_t|\{\boldsymbol{p}\}_{t=1}^{t-1}\right) = P_{sw}\left(\boldsymbol{\varphi}_i|\boldsymbol{A}_\Pi(\{\boldsymbol{p}_t\}_{t=1}^{t-1}))P_i(\boldsymbol{p}_t|\boldsymbol{p}\left[\boldsymbol{\varphi}_i\right]\right)
\end{equation}
where $P_i$ refers to the $i$th component process and $P_{sw}$ refers to the switching process. By recursive application of \eqref{IMP} we obtain:
\begin{equation}
    \label{IMP_joint}
    P(\boldsymbol{p})=P_{sw}\left(\boldsymbol{A_\Pi}(\boldsymbol{p})\right)\prod_{i=1}^M P_i\left(\boldsymbol{p}[\boldsymbol{\varphi}_i]\right)
\end{equation}

\begin{remarks}
  \hfill\par
  \begin{itemize}
    \item The switching process and the component processes are assumed to be ergodic and have unique positive stationary distributions\footnote{Every ergodic Markov chain has a stationary distribution.}. 

    \item   Varying $\boldsymbol{\Pi}$, $\boldsymbol{k}$, the penalized maximum-likelihood cost function can be minimized:
    \begin{equation}
      \label{penalized ML}
      C_{\boldsymbol{\Pi},\boldsymbol{k}}(\boldsymbol{p}) = \hat{H}_{\boldsymbol{\Pi},\boldsymbol{k}}(\boldsymbol{p})+\beta\kappa\log{(n+1)}
    \end{equation}
    where $\hat{H}$ is the empirical entropy of the observed symbol sequence $\boldsymbol{p}$ under an IMP model given partition $\boldsymbol{\Pi}$ and order $\boldsymbol{k}$, and $\beta$ is a non-negative constant, $\kappa$ is defined as:
    \begin{equation}
      \kappa(\boldsymbol{k},\boldsymbol{\Pi}) = \sum_{i=1}^M |\boldsymbol{\varphi}^i|^{k_i}\left(|\boldsymbol{\varphi}^i|-1\right)+(M-1)M^{k_{sw}}
    \end{equation}
    where $|\boldsymbol{\varphi}^i|$ refers to the number of symbols in the corresponding sub-alphabets $\boldsymbol{\varphi}^i$, $k_i$ is the order of the $i$th component chain, $k_{sw}$ is the order of the switching process and $M$ is the number of component chains. In the paper \cite{Zhu2022}, the term $\kappa$ was simplified as the number of parameters that need to be optimized.
    \item The de-interleaving scheme based on penalized maximum-likelihood cost function \eqref{penalized ML} is strongly consistent. In other words, when the data sample $N\to\infty$, the differences between estimated parameters $\hat{\boldsymbol{k}}$, $\hat{\boldsymbol{\Pi}}$ and the original parameters $\boldsymbol{k}$, $\boldsymbol{\Pi}$ tend to be zero.
  \end{itemize}
\end{remarks}

\section{Problem Formulation}
In this section, the generative model for IHMP is proposed, followed by a discussion of the combinatorial problem of the alphabet partitions.

\subsection{The Generative Model}
In the generative model, the hidden states assignments and the parameters are calculated given three pieces of information: the interleaved observation sequence  $\boldsymbol{p}=\{\boldsymbol{p}_t\}_{t=1}^T$, the number of component chain $M$, and the hidden state number of each component chain $\{K^{m}\}_{m=1}^M$. Each component chain can take on $K^{m}$ symbols defined by the corresponding sub-alphabet $\boldsymbol{\varphi}^m$. Each symbol $\boldsymbol{\varphi}^m_k$ is treated as a random variable that follows Gaussian distribution. The generative model of the IHMP is shown in Fig.~\ref{graphical model}(b). Consequently, the joint probability of the generative model is formulated as:
\begin{equation}
  \label{JointProbability}
  \begin{aligned}
      &P(\boldsymbol{Z,S,p,A^z,A,\pi^z,\pi},\mathcal{A}) \propto P(\boldsymbol{Z}_1|\boldsymbol{\pi}^z)\\
      &\prod_{t=2}^T P(\boldsymbol{Z}_t|\boldsymbol{Z}_{t-1},\boldsymbol{A}^z) \times \prod_{m=1}^M\bigg\{P(\boldsymbol{S}_1^m|\boldsymbol{\pi}^m) \\
      &\prod_{t=1}^T P(\boldsymbol{p}_t|\boldsymbol{S}_t^m,Z_{t,m},\mathcal{A}) \prod_{t=2}^T P(\boldsymbol{S}_t^m|\boldsymbol{S}_{t-1}^m,Z_{t,m},\boldsymbol{A}^m) \bigg\}
  \end{aligned}
\end{equation}
where $\boldsymbol{S}_t^m$ is the $K^m \times 1$ state variable associated with the component process, $\boldsymbol{Z}_t$ is the $M \times 1$ state variable associated with the switching process, $\mathcal{A}=\{\boldsymbol{\varphi}_i^m\}$,$i\in[1,K^m]$,$m\in[1,M]$ is the alphabet. The switching process determines which of the component chains is active while the other chains remain idle (the component chain stays in the previous state). Specifically:
\begin{equation}
  \label{transition}
  \begin{aligned}
  P(\boldsymbol{S}_t^m=i|\boldsymbol{S}_{t-1}^m=j,\boldsymbol{Z}_t=k)=
    \begin{cases}
      {A}^m_{j,i},& k=m;\\
      {E}^m_{j,i},& k\neq m,
    \end{cases}&\\
    =({{A}_{j,i}^m})^{Z_{t,m}}({{E}_{j,i}^m})^{(1-Z_{t,m})}&
  \end{aligned}
\end{equation}
where $\boldsymbol{\pi}^m$, $ \boldsymbol{A}^m$ are the prior distribution and the transition matrix of the $m$th component process, $\boldsymbol{\pi}^z$, $ \boldsymbol{A}^z$ is the prior distribution and transition matrix of the switching process, and $\boldsymbol{E}^m$ is the identity matrix with size $K^m$. Here, we treat the symbol in the alphabet as a random variable, and $\mathcal{A}$ is the set of Gaussian distributions wherein $\boldsymbol{\varphi}_i^m\sim\mathcal{N}(\boldsymbol{\mu}_i^m,\boldsymbol{\Sigma}_i^m)$. Thus, the observation follows the multivariate Gaussian distribution defined by mean and covariance:
\begin{equation}
  \label{GaussianDistribution}
  \begin{aligned}
      &P(\boldsymbol{p}_t|\boldsymbol{S}_t^m=i,\boldsymbol{Z}_{t,m}=m,\mathcal{A})=f_{\boldsymbol{\varphi}_i^m}(\boldsymbol{p}_t)=\\
      &\frac{1}{{C}^m}\exp\bigg(-\frac{1}{2}(\boldsymbol{p}_t-\boldsymbol{\mu}_i^m)^\top(\boldsymbol{\Sigma}_i^m)^{-1}(\boldsymbol{p}_t-\boldsymbol{\mu}_i^m)\bigg)
  \end{aligned}
\end{equation}
where $C^m$ is the normalization term.

In the joint probability function \eqref{JointProbability}, the prior distribution terms $P(\boldsymbol{A}^m)$, $P(\boldsymbol{A}^z)$, $P(\boldsymbol{\pi}^m)$, $P(\boldsymbol{\pi}^z)$, and $P(\mathcal{A})$ are canceled for the simplicity of representation\footnote{In other words, the prior distributions are random variables follow the uniform distribution. The conjugate prior can be easily added to \eqref{JointProbability} according to \cite{Bishop2006}.}.
The objective is to infer the hidden states and estimate the parameters of the generative model based on the observations. The objective can be formulated via the Bayesian theorem:
\begin{equation}
  P(\boldsymbol{Z},\boldsymbol{S},\boldsymbol{\Gamma}|\boldsymbol{p}) = \frac{P(\boldsymbol{Z},\boldsymbol{S},\boldsymbol{\Gamma})P(\boldsymbol{p}|\boldsymbol{Z},\boldsymbol{S},\boldsymbol{\Gamma})}{\int P(\boldsymbol{p},\boldsymbol{Z},\boldsymbol{S},\boldsymbol{\Gamma})d (\boldsymbol{Z},\boldsymbol{S},\boldsymbol{\Gamma})}
\end{equation}
where $\boldsymbol{\Gamma}=\{\boldsymbol{A},\boldsymbol{A}^z,\boldsymbol{\pi},\boldsymbol{\pi}^z,\mathcal{A}\}$, and $P(\boldsymbol{Z},\boldsymbol{S},\boldsymbol{\Gamma})$ is the prior distribution; $\int P(\boldsymbol{p}^t,\boldsymbol{Z},\boldsymbol{S},\boldsymbol{\Gamma})d (\boldsymbol{Z},\boldsymbol{S},\boldsymbol{\Gamma})$ is the evidence integral. We infer the hidden states and estimate the parameters by maximizing the posterior:
\begin{equation}
  (\boldsymbol{Z}^*,\boldsymbol{S}^*,\boldsymbol{\Gamma}^*) = \arg\max_{\boldsymbol{Z,S,\Gamma}} P({\boldsymbol{Z},\boldsymbol{S},\boldsymbol{\Gamma}}|\boldsymbol{p})
\end{equation}
where $\boldsymbol{Z}^*$,  $\boldsymbol{S}^*$, and  $\boldsymbol{\Gamma}^*$ refer to the optimal solution of hidden variables.

\subsection{The Combinatorial Problem of IMP and IHMP}
In the context of IMP \cite{Seroussi2012}, the de-interleaving objective is to infer the alphabet partition $\boldsymbol{\Pi}$ and the order of Markov chains $\boldsymbol{k} = (k_1, k_2,..., k_m, k_w)$, which is optimized by exhaustive search or heuristic search. On the one hand, the order vector is searched according to the Bayesian Information Criterion (BIC) \cite{Csiszar2000}, which is convenient to implement. On the other hand, the major computational burden arises from searching the alphabet partitions. The partition searching space has $\Omega_\mathcal{A}(|\mathcal{A}|)=\sum_i^{|\mathcal{A}|} Stirling(\mathcal{|\mathcal{A}|},i)$ cases, wherein $Stirling$ is the Stirling number of the second kind \cite{graham1989concrete}. The search space size grows exponentially as the size of the alphabet increases, which is impractical for real-life applications.

In terms of IHMP, we first simplify the problem by limiting the order of each Markov chain to one since it is enough for modeling real-life time-series, e.g., radar signal modeling \cite{Bao2023,Zhu2021} or aircraft recognition \cite{Du2011bayesian}. Secondly, we extend the component chains from Markov to hidden Markov. Following these two steps, the learning and inference problem has three variables to be optimized: the alphabet partition $\boldsymbol{\Pi}$, the hidden states assignments $\boldsymbol{Z}$, $\boldsymbol{S}$, and the model parameters $\boldsymbol{\Gamma}$. With the generative model described in the previous part, $\boldsymbol{Z}$, $\boldsymbol{S}$, $\boldsymbol{\Gamma}$ is optimized given the number of component chains $M$ and the hidden state number of each component chain $K^m$. Thus, the search space of alphabet partition has $\Omega_\mathcal{A}^\prime (N) = \sum_{i=1}^{N} Partition(N,i)$ cases, where $Partition$ is the partition number \cite{ewell2004recurrences} in the number theory. The partition number is much smaller than the sum of the Stirling number of the second kind. e.g., $\Omega_\mathcal{A}^\prime(10)=42$ and $\Omega_\mathcal{A}(10)=115975$. 

In general, introducing IHMP makes the search space much smaller than the previous IMP. In the next section, we propose efficient algorithms to perform the posterior inference of the generative model.

\section{Method}
In a probabilistic model, the inference problem involves computing the probability of the hidden states given the observations. The learning problem for a probabilistic graphical model consists of two components: learning the structure of the model and its parameters. We only learn the model parameter in this paper. This approach is achieved via the Viterbi algorithm \cite{Viterbi1967} or its variation \cite{Landwehr2008}. Viterbi-like algorithm is a form of dynamic programming that is very closely related to the forward-backward algorithm. However, Viterbi-like algorithms require the model parameter as its input. In this paper, we perform inference and learning at the same time by designing EM-like algorithms. 

In this section, firstly, we proposed an exact inference method based on the EM algorithm in the spirit of \cite{Minot2014}. However, exact inference is shown to be NP-hard. Secondly, we proposed an approximate inference method based on variational inference, two variational distributions were designed to approximate the posterior. Note that for the simplicity of representation and illustration, the covariance matrix of each symbol is assumed to be equal ($\boldsymbol{\Sigma}_i^m=\boldsymbol{\Sigma}, i \in [1,K^m], m \in [1,M]$).

\subsection{Exact Inference}
The exact EM algorithm is a general technique for finding maximum likelihood solutions for probabilistic graphical models having hidden variables \cite{Bishop2006}. The EM algorithm iteratively computes the expectation of the complete data log-likelihood (E-step) and optimizes its parameters (M-step). However, the E-step is computationally intractable. This fact is analyzed as follows.
To perform exact inference, we transform the generative model in Fig. \ref{graphical model}(b) into an equivalent Markov chain with $M\prod_{m=1}^{M}K^m$ hidden states and use the standard forward-backward algorithm to perform E-step. The exact algorithm is performed in space $O(TM\prod_{m=1}^{M}K^m)$ with time $O(TM^2\prod_{m=1}^{M}(K^m)^2)$. This time complexity includes the term of the product of the component chain hidden state number, the exponential time complexity makes the exact E-step intractable.

\begin{theorem}
    Exact inference (E-step) for interleaved mixtures of hidden Markov models is NP-hard~\cite{Landwehr2008}.
\end{theorem}

The EM algorithm follows from the definition of the expected joint log-likelihood of the hidden and observed variables:
\begin{equation}
  \label{EMq}
  \mathbb{Q}(\boldsymbol{\Gamma}^\prime|\boldsymbol{\Gamma}) = \mathbb{E}\left\{\log P (\boldsymbol{S},\boldsymbol{Z},\boldsymbol{p}|\boldsymbol{\Gamma}^\prime)|\boldsymbol{\Gamma},\boldsymbol{p}\right\}
\end{equation}
where $\boldsymbol{\Gamma}^\prime$ is the new parameter updated in an iteration. The E-step involves calculating the $\mathbb{Q}$ function \eqref{EMq} and the M-step consists of updating the parameter $\boldsymbol{\Gamma}$. Specifically, the update function is obtained by solving the combination of \eqref{JointProbability}, \eqref{transition}, \eqref{GaussianDistribution}. The $\mathbb{Q}$ function is rewritten as:
\begin{equation}
  \label{Qfunction}
  \begin{aligned}
    \mathbb{Q} &=\sum_{m=1}^M {\mathbb{E}(\boldsymbol{S}_1^m)}^\top \log \boldsymbol{\pi}^m +\mathbb{E}(\boldsymbol{Z}_t)^\top \log \boldsymbol{\pi}^z\\
    &+\sum_{t=2}^T \bigg\{\sum_{m=1}^M {\mathbb{E}(\boldsymbol{S}_t^m)}^\top \mathbb{E}(\bar{\boldsymbol{A}}_t^m) \mathbb{E}(\boldsymbol{S}_{t-1}^m)\\
    &+{\mathbb{E}(\boldsymbol{Z}_t)}^\top \log \boldsymbol{A}^z \mathbb{E}(\boldsymbol{Z}_{t-1})\bigg\} \\
    &-\frac{1}{2}\sum_{t=1}^T\Bigg[\left(\boldsymbol{p}_t-\sum_{m=1}^M \mathbb{E}(Z_{t,m}{\boldsymbol{S}_t^m}^\top)\boldsymbol{\mu}^m\right)\boldsymbol{\Sigma}^{-1} \\
    &\ \ \ \ \ \ \ \ \ \ \ \ \left(\boldsymbol{p}_t-\sum_{m=1}^M \mathbb{E}({Z_{t,m}{\boldsymbol{S}_t^m}^\top})\boldsymbol{\mu}^m\right)^\top\Bigg]-C^m
  \end{aligned}
\end{equation}
where $\bar{\boldsymbol{A}}_t^m = (1-Z_{t,m})\log \boldsymbol{E}^m + Z_{t,m} \log \boldsymbol{A}^m$, $\mathbb{E}(\cdot)$ means taking expectation with respect to the hidden variables,  $\boldsymbol{\mu}^m$ is a mean vector consisting of $\boldsymbol{\mu}^m_i$, $i\in[1, K^m]$ and $C^m$ arises from the normalization term. The M-step for estimating parameters is obtained by setting the derivative of $\mathbb{Q}$ equal to zero. The details of the M-step can be found in Appendix A.

\subsection{Approximate Inference}

Approximate inference in the probabilistic graphical model has caught much attention \cite{Zhang2018}. The Markov Chain Monte Carlo (MCMC) algorithm like blocked Gibbs sampling with forward filtering and backward sampling can potentially solve the problem in the state space model \cite{Fox2011}. However, it is well-known that the MCMC-based algorithm is time-consuming since the sampling process is effective when the Markov chain is burn-in~\cite{Murphy2012}. While it has a theoretical guarantee to converge to true posterior, the time-consuming nature makes it not suitable for many signal processing applications like radar signal processing \cite{Zhu2021}, multi-target sensing \cite{valera2015infinite}, tracking\cite{vo2016efficient}, etc. Alternatively, variational inference provides an efficient solution for model inference, which can also be easily extended to the online paradigm \cite{broderick2013streaming}.

\begin{figure}[!t]
  \centering
  \includegraphics[width=3.4in]{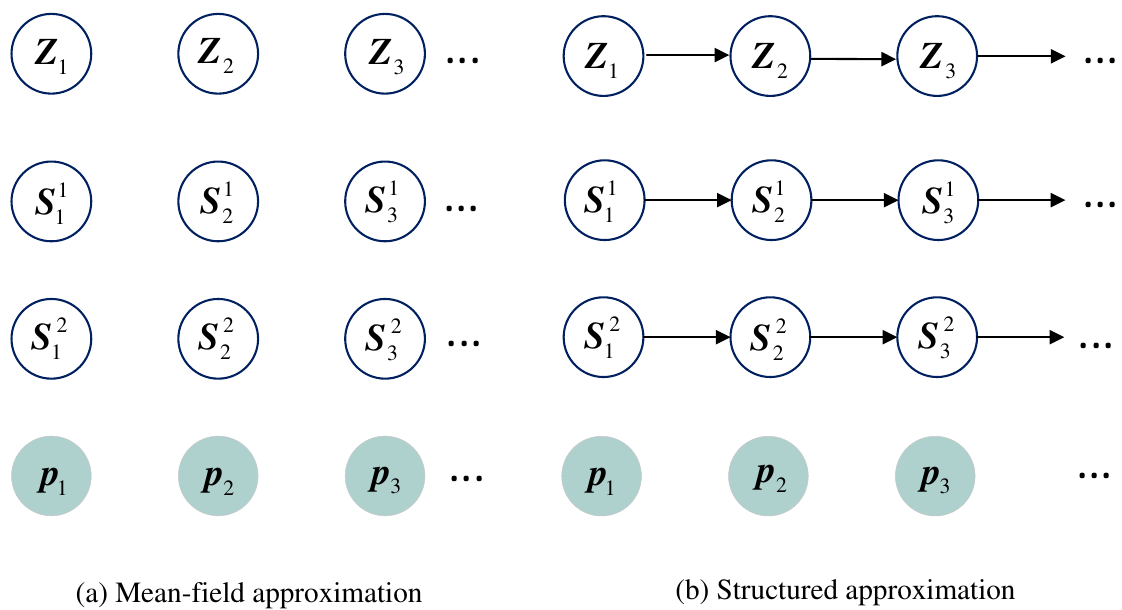}
  \caption{The mean-field approximation and structured approximation.}
  \label{variational model}
  \end{figure}

The essence of the variational inference is to introduce a variational distribution $Q$ and minimize the distance between the true posterior $P$ and variational distribution $Q$. The log-likelihood function of the observed sequence is formulated as:
\begin{equation}
  \begin{aligned}
    &\ln P(\boldsymbol{p}|\boldsymbol{Z, S, \Gamma}) =\\
    &\mathcal{L}(Q(\boldsymbol{Z},\boldsymbol{S},\boldsymbol{\Gamma}))+ \mathcal{KL}(Q(\boldsymbol{Z, S, \Gamma})||P(\boldsymbol{Z,S,\Gamma|p}))
  \end{aligned}
\end{equation}
where $\mathcal{L}$ denotes the well-known evidence lower bound (ELBO), and $\mathcal{KL}$ is the Kullback-Leibler divergence. Minimizing the $\mathcal{KL}$ between the variational distribution and the exact posterior is equivalent to maximizing the ELBO due to Jensen's inequality [36]. Specifically, the ELBO is formulated as:

\begin{equation}
  \label{lowerbound}
  \mathcal{L} = \mathbb{E}_Q\{\ln P(\boldsymbol{\Gamma,S,Z,p})\} - \mathbb{E}_Q\{\ln Q(\boldsymbol{S,Z,\Gamma})\}
\end{equation}
where $\mathbb{E}_Q$ means taking expectation with respect to variational distribution $Q$. According to Blei~\cite{Blei2017}, the complexity of variational inference is determined by the conditional independence relations. Thus, tractable variational structures need to be designed. Given the variational structure, the variational parameters need to be optimized to obtain the tightest bound by maximizing the ELBO \eqref{lowerbound}.  In this paper, we have derived two variational inference methods explicitly based on the mean-field assumption and the structured assumption.

\subsubsection{Mean-field Variational Inference (MFVI)}
A simple choice is the mean-field assumption, shown in Fig.~\ref{variational model}(a). The likelihood is formulated as:
\begin{equation}
  \label{MF}
    Q(\boldsymbol{S,Z,\Gamma}) =  Q(\boldsymbol{\Gamma})\prod_{t=1}^T \prod_{m=1}^M Q(\boldsymbol{S}_t^m|\boldsymbol{\theta}_t^m) \prod_{t=1}^T Q(\boldsymbol{Z}_t|\boldsymbol{\phi}_t)
\end{equation}
where $\boldsymbol{\theta}_t^m$ and $\boldsymbol{\phi}_t$ are the expectations of the state variables of the $m$th component chain and switching chain at time $t$; $\boldsymbol{S}_t^m$, $\boldsymbol{Z}_t$ are state variables with 1 in the $k$th position and 0 elsewhere. Given the above notations, we can explicitly express the state variables as $Q(\boldsymbol{S}_t^m|\boldsymbol{\theta}_t^m) = \prod_{k=1}^{K^m} \big(\theta_{t,k}^m \big)^{S_{t,k}^m}$ and $Q(\boldsymbol{Z}_t|\boldsymbol{\phi}_t) = \prod_{m=1}^M \big(\phi_{t,m} \big)^{Z_{t,m}}$, where the first term of the subscript $t$ is the time index, and the second term of the subscript $m$ is the element index of the vector. We set $Q(\boldsymbol{\Gamma}) = 1$. The ELBO is rewritten as:

\begin{equation}
  \label{MF_ELBO}
  \begin{aligned}
    \mathcal{L} = &-\sum_{t=1}^{T} \sum_{m=1}^{M} {\boldsymbol{\theta}_t^m}^\top \log \boldsymbol{\theta}_t^m-
    \sum_{t=1}^{T} {\boldsymbol{\phi}_t}^\top \log \boldsymbol{\phi}_t\\
    &-\frac{1}{2}\sum_{t=1}^T\bigg[\boldsymbol{p}_t\boldsymbol{\Sigma}^{-1}{\boldsymbol{p}_t}^\top
-2\sum_{m=1}^M \boldsymbol{p}_t \boldsymbol{\Sigma}^{-1}{\boldsymbol{\mu}^m}^\top\phi_{t,m}\boldsymbol{\theta}_t^m \\
    &+\sum_{m=1}^M tr\{\boldsymbol{\Delta}_m^m diag(\boldsymbol{\phi}_{t,m}\boldsymbol{\theta}_t^m)\}\\
    &+\sum_{m=1}^{M} \sum_{\substack{n=1\\n\neq m}}^{M} tr \{\boldsymbol{\Delta}_m^n (\phi_{t,n}\boldsymbol{\theta}_t^n)(\phi_{t,m} \boldsymbol{\theta}_t^m)^\top \}
    \bigg]\\
    &+\sum_{m=1}^{M} {\boldsymbol{\theta}_1^m}^\top \log \boldsymbol{\pi}^m
    + \sum_{t=2}^{T} \sum_{m=1}^{M} {\boldsymbol{\theta}_t^m}^\top \mathbb{E}_Q(\bar{\boldsymbol{A}}_t^m) \boldsymbol{\theta}_{t-1}^m\\
    &+\boldsymbol{\phi}_1^\top \log \boldsymbol{\pi}^z+\sum_{t=2}^T\boldsymbol{\phi}_t^\top \boldsymbol{A}^z \boldsymbol{\phi}_t
  \end{aligned}
\end{equation}
where $\boldsymbol{\Delta}_m^n = \boldsymbol{\mu}^m \boldsymbol{\Sigma}^{-1}{\boldsymbol{\mu}^n}^\top$ and $diag$ is an operator that takes a vector and returns a square matrix with the elements of the vector along its diagonal, The tightest bound is obtained by taking the derivative of \eqref{MF_ELBO}, and setting it to zero. The update function of the hidden states is shown in \eqref{mfupdatetheta} and \eqref{mfupdatephi}. The derivation of the ELBO and the update function are shown in Appendix B. In \eqref{mfupdatetheta} and \eqref{mfupdatephi}, $idiag$ is an operator that takes a square matrix and returns a vector along its diagonal. Note that the fourth and the fifth terms of \eqref{mfupdatetheta} arise from the correlation of the state variable sequence. Although the posterior distribution over the hidden variables is approximated by the mean-field assumption, the time-series dependencies are still retained. The time dependencies propagate information along the same pathways as those defining the exact algorithms for probability propagation. Each hidden state vector is updated using \eqref{mfupdatetheta} and \eqref{mfupdatephi} with time complexity $O(T\sum_{m=1}^{M}(K^m)^2+TM^2)$ in space $O(T\sum_{m=1}^{M}K^m+MT)$ per iteration.
\begin{figure*}
  \begin{equation}
    \label{mfupdatetheta}
    \boldsymbol{\theta}_t^m = \exp \bigg\{
      \boldsymbol{\mu}^m \boldsymbol{\Sigma}^{-1}  \boldsymbol{p}_t \phi_{t,m}-\frac{1}{2} \sum_{\substack{n=1\\n\neq m}}\boldsymbol{\Delta}_m^n (\phi_{t,n}\boldsymbol{\theta}_t^n) \phi_{t,m}-\frac{1}{2}idiag\{\boldsymbol{\Delta}_m^m\phi_{t,m}\} + \mathbb{E}_Q({\bar{\boldsymbol{A}}_t^m}) \boldsymbol{\theta}_{t-1}^m + \mathbb{E}_Q(\bar{\boldsymbol{A}}_t^m )\boldsymbol{\theta}_{t+1}^m
    \bigg\}
  \end{equation}
  \begin{equation}
    \label{mfupdatephi}
    \phi_{t,m} = \exp \bigg\{
  \boldsymbol{p}_t\boldsymbol{\Sigma}^{-1}{\boldsymbol{\mu}^m}^\top \boldsymbol{\theta}_t^m + \frac{1}{2}tr\left(\boldsymbol{\Delta}_m^m diag(\boldsymbol{\theta}_t^m)\right) + \log A_{m,m}^z + \sum_{\substack{n=1\\n\neq m}}^M \phi_{t,n}\log A_{m,n}^z+\frac{1}{2}tr\left\{\boldsymbol{\Delta}_m^n\phi_{t,n}\boldsymbol{\theta}_t^n{\boldsymbol{\theta}_t^m}^\top\right\}
    \bigg\}
  \end{equation}
\end{figure*}

\subsubsection{Structured Variational Inference (SVI)}
The MFVI takes the relatively extreme assumption. Alternatively, we can preserve the model structure while making it tractable. Following such requirements, we preserve the horizontal couplings, as illustrated in Fig. \ref{variational model}(b), the E-step of $M+1$ Markov chains is performed via an efficient forward-backward algorithm. The variational likelihood is formulated as:

\begin{equation}
  \label{Structured variational family}
\begin{aligned}
     Q(\boldsymbol{S,Z,\Gamma}) = & Q(\boldsymbol{\Gamma})\prod_{m=1}^M\bigg\{Q(\boldsymbol{S}_1^m)\prod_{t=2}^T Q(\boldsymbol{S}_t^m|\boldsymbol{S}_{t-1}^m)\bigg\}\\
     \times &Q(\boldsymbol{Z}_1)\prod_{t=2}^T Q(\boldsymbol{Z}_t|\boldsymbol{Z}_{t-1})
\end{aligned}
\end{equation}
where $Q(\boldsymbol{S}_1^m)=\prod_{k=1}^{K^m} ({h}_{1,k}^m \pi_k^m)^{S_{1,k}^m}$ is the initial density and $Q(\boldsymbol{S}_t^m|\boldsymbol{S}_{t-1}^m) = \prod_{j=1}^{K^m} (\prod_{i=1}^{K}({h}_{t,i}^mA_{j,i}^m)^{S_{t,i}^m})^{S_{t-1,j}^m}$ is the transition probability of the $m$th component chain. Meanwhile $Q(\boldsymbol{Z}_1) = \prod_{k=1}^{M} g_{1,k}\pi_{1,k}^z$ is the initial density, and  $Q(\boldsymbol{Z}_t|\boldsymbol{Z}_{t-1}) = \prod_{j=1}^M(\prod_{i=1}^{M}(g_{t,i}A_{j,i}^z)^{Z_{t,i}})^{Z_{t-1,j}}$ is the transition probability of the switching chain. The lower bound is formulated as:

  
\begin{align}
    \label{structureLowerBound}
    \mathcal{L} =& -\sum_{t=1}^{T} \sum_{m=1}^{M} {\boldsymbol{\theta}_t^m}^\top \log \boldsymbol{h}_t^m-
    \sum_{t=1}^{T} {\boldsymbol{\phi}_t}^\top \log \boldsymbol{g}_t \nonumber \\
    &-\frac{1}{2}\sum_{t=1}^T\bigg[\boldsymbol{p}_t^\top\boldsymbol{\Sigma}^{-1}{\boldsymbol{p}_t}-2\sum_{m=1}^M\boldsymbol{p}_t \boldsymbol{\Sigma}^{-1}{\boldsymbol{\mu}^m}^\top\phi_{t,m}\boldsymbol{\theta}_t^m \nonumber \\
    &-\sum_{m=1}^M tr\{\boldsymbol{\Delta}_m^m diag\{\phi_{t,m}\theta_t^m\}\}\\
    &+\sum_{m=1}^M\sum_{\substack{n=1\\n\neq m}}^{M} tr \{\boldsymbol{\Delta}_m^n (\phi_{t,n} \boldsymbol{\theta}_t^n) (\phi_{t,m}\boldsymbol{\theta}_t^m)^\top\}
    \bigg] \nonumber \\
    &+\sum_{t=2}^{T} \sum_{m=1}^{M} (1-\phi_{t,m}) {\boldsymbol{\theta}_t^m}^\top (\log \boldsymbol{A}^m-\log \boldsymbol{E}^m) \boldsymbol{\theta}_{t-1}^m \nonumber
\end{align}

Similar to the mean-field variational inference, the update functions \eqref{structureupdatetheta} and \eqref{structureupdatephi} are obtained by taking the derivative with respect to $\boldsymbol{\theta}$ and $\boldsymbol{\phi}$ and setting it to zero. The update functions derived in Appendix C. Note that $\boldsymbol{\Gamma}$ remains equal to the equivalent parameters of the true system, i.e., $P(\boldsymbol{\Gamma}) = Q(\boldsymbol{\Gamma})$. Intuitively, SVI uncouples the $M+1$ Markov chains and attaches to each state variable a distinct observation.
\begin{figure*}
  \begin{equation}
    \label{structureupdatetheta}
    \boldsymbol{h}_t^m = \exp \bigg\{
      \boldsymbol{\mu}^m\ \boldsymbol{\Sigma}^{-1} {\boldsymbol{p}_t}^\top - \frac{1}{2}idiag\{\boldsymbol{\Delta}_m^m\phi_{t,m}\} - (1-\phi_{t,m})(\boldsymbol{\theta}_{t-1}^m+\boldsymbol{\theta}_{t+1}^m) (\log \boldsymbol{E}^m - \log \boldsymbol{A}^m) -\frac{1}{2}\sum_{\substack{n=1\\n\neq m}}^M \boldsymbol{\Delta}_m^n (\phi_{t,n}\boldsymbol{\theta}_t^n)\phi_{t,m}
    \bigg\}
  \end{equation}
  \begin{equation}
    \label{structureupdatephi}
    g_{t,m} = \exp \bigg\{
      \boldsymbol{p}_t\boldsymbol{\Sigma}^{-1}{\boldsymbol{\mu}^m}^\top \boldsymbol{\theta}_t^m - \frac{1}{2}tr\{\boldsymbol{\Delta}_m^m diag(\boldsymbol{\theta}_t^m) \}- \sum_{\substack{n=1\\n\neq m}}^M tr\{\boldsymbol{\Delta}_m^n\phi_{t,n}\boldsymbol{\theta}_t^n{\boldsymbol{\theta}_t^m}^\top\}-{\boldsymbol{\theta}_t^m}^\top(\log \boldsymbol{E}^m-\log \boldsymbol{A}^m)\boldsymbol{\theta}_{t-1}^m
    \bigg\}
  \end{equation}
\end{figure*}

In \eqref{structureupdatetheta} and \eqref{structureupdatephi}, $\boldsymbol{\phi}$ and $\boldsymbol{\theta}$ are the expectations of the hidden variables corresponding to the switching process and component processes. $\boldsymbol{g}$ and $\boldsymbol{h}$ are the emission variables corresponding to the switching process and component processes. Using the emission probabilities and the forward-backward algorithm to calculate the likelihood. The time complexity of the structure is $O(MTK^2+TM^2)$ in space $O(MKT+MT)$ per iteration.

\section{Error Analysis}
In this section, we develop a lower bound on the error probability for de-interleaving HMMs with Gaussian emission based on the likelihood ratio test. The bound is used to evaluate how close an algorithm is to the optimum. We derived the error probability lower bound when $M=2$, $K^1=K^2=2$. We then compare the error probability performance on various algorithms and the lower bound proposed in Theorem 2. Note that, the lower bound is derived for disjoint sub-alphabets in this research, the non-disjoint sub-alphabets counterpart needs further investigation.

\subsection{Theoretical Error Lower Bound}
Let $p_i$ be the $i$th observation and $d_i\in\{1,2\}$ be the decision of the $i$th pulse. We modeled the component process by a first-order Markov chain. The optimum decision $\hat{d}_i$ can be formulated when $d_{i-1}$ is fixed:
\begin{equation}
  \label{optimumdecision}
  \begin{aligned}
      \hat{d}_i &= \arg \max P(d_i|p_i,d_{i-1}) \\
      &=\arg \max\frac{P(p_i|d_i,d_{i-1})P(d_i|d_{i-1})}{P(p_i)}
  \end{aligned}
\end{equation}

In general, two sources can be confused in many ways, we consider one kind of error event since we are developing a lower bound. Considering an error event $\mathcal{E}_i^{xy}$ where the $i$th observation is actually from source $y$, but it is identified to be from another source. Meanwhile, the previous observation is generated from source $x$. The error probability is formulated as follows:
\begin{equation}
  \label{errorprobability}
  \begin{aligned}
    P_{e}&\geq \lim_{n\to \infty} \mathbb{E}\bigg[\frac{1}{n}\sum_{i=1}^{n} \sum_{x=1}^2\sum_{y=1}^2 \xi^z_x{A}_{x,y}^z\mathbb{I}(\mathcal{E}_i^{xy})\bigg]\\
    &=\sum_{x=1}^2\sum_{y=1}^2 \xi_x^z{A}_{x,y}^z P(\mathcal{E}_i^{xy})
  \end{aligned}
\end{equation}
where $A^z_{x,y}$ refers to the switching chain's transition probability from source $x$ to source $y$. 
\begin{theorem}
   There are two binary-state ergodic HMMs with Gaussian emissions. The stationary distribution of each component chain is $\Xi^m=[\xi_1^m, \xi_2^m]$, and the mean and variance of the Gaussian emissions of a stationary Markov chain are $[(\mu_1^i,\sigma_1^i),(\mu_2^i,\sigma_2^i)]$. The transition matrix of the switching chain is $\boldsymbol{A}^z$. The transition matrix of the component chain is $\boldsymbol{A}^m$. Suppose the sub-alphabets are disjoint and $\mu_1^i<\mu_2^i$, $\sigma_1^i=\sigma_2^i=\sigma$. The error probability is bounded by: 
    \begin{equation}
     \label{lower bound}
        \begin{aligned}
            P_e \geq &\sum_{x=1}^2\sum_{y=1}^2 \bigg\{\xi_x^z{A}_{x,y}^z  \times\\
            &  \sum_{k=1}^2 \sum_{l=1}^2 \xi_k^y\xi_l^{\backslash y }\mathcal{Q}\left(\frac{(-1)^{u(\mu_k^y-\mu_l^{\backslash y})}(\gamma_{kl}^{xy}-\mu_l^r)}{\sigma}\right)\bigg\}
        \end{aligned}
    \end{equation}
where $\mathcal{Q}$ is the right-tail function and $\sigma_1^i = \sigma_2^i = \sigma$, $u(\cdot)$ is the unit function, $\gamma_{kl}^{xy}$ is illustrated as follows:
\begin{equation}
\begin{aligned}
     &\gamma_{kl}^{xy}=\frac{({\mu_k^y})^2-({\mu_l^{\backslash y}})^2}{2(\mu_k^y-\mu_l^{\backslash y})}+\\
     &2\sigma^2\frac{\log\frac{\xi_k^y {A}^z_{x,y}}{\xi_k^{\backslash y} {A}_{x,{\backslash y}}^z}+{\Xi^y}^\top\log \boldsymbol{A}_{\cdot,k}^y-{\Xi^{\backslash y}}^\top\log \boldsymbol{A}_{\cdot,l}^{\backslash y}}{2(\mu_k^y\xi_k^y-\mu_l^{\backslash y}\xi_l^{\backslash y})}
\end{aligned}
\end{equation}
where $\backslash{y}=1$ if $y=2$, $\backslash y=2$  if $y=1$, and $\boldsymbol{A}_{\cdot,l}^m$ is the $l$th column of the transition matrix $\boldsymbol{A}$ of the $m$th component chain. 
\end{theorem}

The proof can be found in Appendix D.


\subsection{Performance Comparison}
\begin{figure*}[!t]
  \centering
  \includegraphics[width=6.8in]{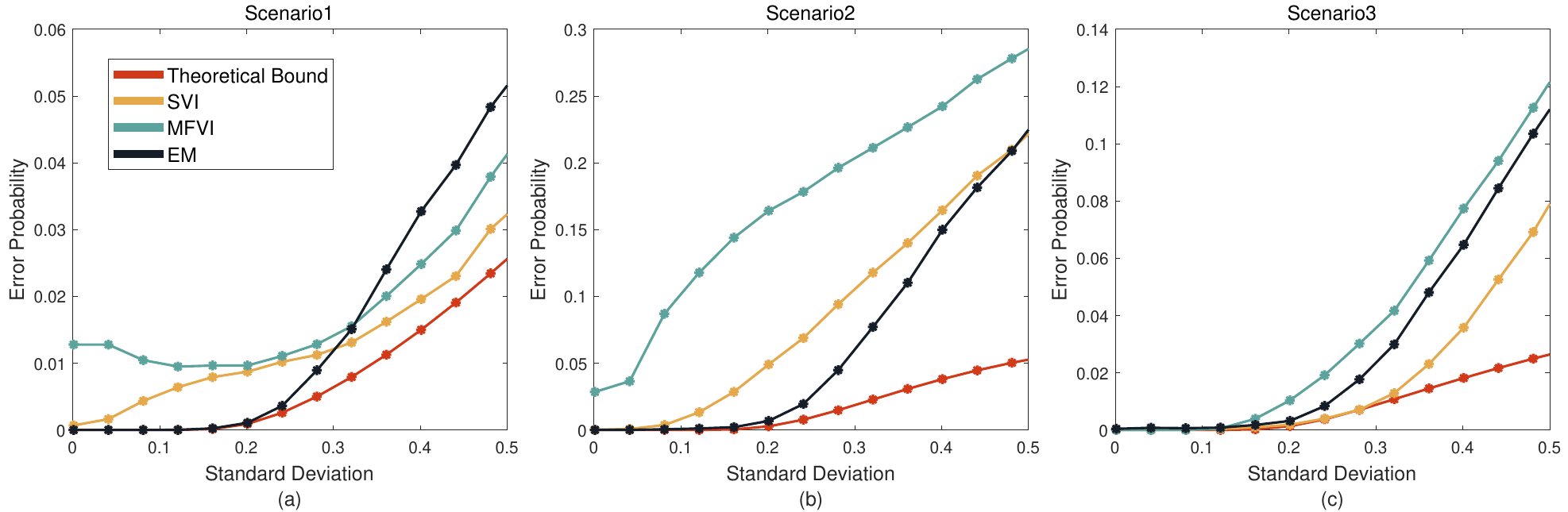}
  \caption{The error probability of three scenarios was averaged over 100 trials. The results are compared with the theoretical lower bound \eqref{lower bound} shown as the red line. Scenario 1, scenario 2, scenario 3 corresponds to $[\mu_1^1,\mu_2^1, \mu_1^2, \mu_2^2] \in \{[1,2,3,4],[1,3,2,4],[1,3,4,2]\}$.}
  \label{error analysis}
\end{figure*}

In this part, we compare the three proposed algorithms, as well as the lower bound described in Theorem 2. Specifically, we set two hidden Markov models with transition probability  $\boldsymbol{A}^1=\boldsymbol{A}^2=[0.1,0.9;0.9,0.1]$. The emission function of each symbol $\boldsymbol{\varphi}_i^m$ follows $\mathcal{N}(\mu_i^m, \sigma^2)$. We set up three different scenarios using different means $\mu_i^m$, and three scenarios are $\mu_1^1,\mu_2^1, \mu_1^2, \mu_2^2 \in \{[1,2,3,4],[1,3,2,4],[1,3,4,2]\}$\footnote{The problem of separating two binary Markov processes reduce to these three scenarios. Analysis can be found in Appendix D.}. The switching chain has transition density  $\boldsymbol{A}^z=[0.1,0.9;0.9,0.1]$. We run these three chains until stationary. Then $N=900$ interleaved observations were intercepted to perform further experiments. We compute the de-interleaving error by:
\begin{equation}
  P_e = \frac{\sum_i\mathbbm{I}(\hat{d_i}\neq d_i)}{N}
\end{equation}
where $\mathbbm{I}$ is the indicator function. We adjust the Standard Deviation (SD) of each symbol from $0$ to $0.5$ with step $0.1$. The average error probabilities over the 100 trials are computed and compared with the lower bound. Results are shown in Fig.~\ref{error analysis}. 

All of the proposed algorithms have effective performance. Specifically, the EM algorithm is the closest to the lower bound when the standard deviation is low in scenarios 1 and 2. In scenario 3, the SVI is the closest to the lower bound. The larger the standard deviation, the higher the error probability of the algorithms and the bounds. Note that the variational approach may not always approach zero when the standard deviation approaches zero, whereas exact EM does. This phenomenon indicated that the approximate algorithms and exact algorithms lack consistency when the standard deviation approaches zero. Meanwhile, it is already proven that the mean-field approximation lacks consistency under a large sample limit \cite{Wang2004} for the state space model. But naturally, with some confidence, we can substitute these variational methods for exact ones thanks to their low complexity and error  probability. What's more, compared with the MFVI, the SVI has a lower error probability in most cases.

\section{Simulations}
We now describe a Monte Carlo simulation to investigate the performance of MFVI, SVI, and EM algorithms for de-interleaving mixtures of HMPs. Firstly, we describe the simulation scenarios with disjoint sub-alphabets and non-disjoint sub-alphabets. Secondly, the de-interleaving performance of these two scenarios is quantified by de-interleaving accuracy and mean square error. Finally, the robustness under missing observations is tested.
\subsection{Data Description}
There are three sources and consistently emit signals. The detail of each source is described in Table.~\ref{data description}. The symbols are defined by ($\mu,\sigma^2$) pairs corresponding to independent draws from a Gaussian distribution with mean $\mu$ and variance $\sigma^2$. The transition density of component chains and switching chains are set as:
\begin{equation}
    \boldsymbol{A}^1=\boldsymbol{A}^2=\boldsymbol{A}^3=\boldsymbol{A}^z=\left( 
    \begin{array}{cc}
        \beta_1 & 1-\beta_1 \\
        1-\beta_2 & \beta_2 
    \end{array}
    \right)
\end{equation}
where $\beta_1=\beta_2=0.1$ are set. 

\begin{table}
  \caption{Source Description for Interleaved Scenario}
\centering
  \label{data description}
  \begin{tabular}{cccl}
    \hline
    \makecell[c]{HMP \\Index} & Symbol 1 & Symbol 2\\
    \hline
    \multicolumn{3}{c}{Disjoint Sub-alphabets}\\
    \hline
    1 & $(1,\sigma^2)$ & $(2,\sigma^2)$\\
    2 & $(4,\sigma^2)$ & $(5,\sigma^2)$\\
    3 & $(7,\sigma^2)$ & $(8,\sigma^2)$\\
    \hline
    \multicolumn{3}{c}{Non-disjoint Sub-alphabets}\\
    \hline
    1 & $(1,\sigma^2)$ & $(4,\sigma^2)$\\
    2 & $(4,\sigma^2)$ & $(5,\sigma^2)$\\
    3 & $(7,\sigma^2)$ & $(8,\sigma^2)$\\
    \hline
  \end{tabular}
\end{table}

\subsection{Evaluation Metrics}
We evaluate the performance by de-interleaving accuracy and parameter estimation Mean Square Error (MSE):

\textbf{Accuracy}: The de-interleaving accuracy reflects the ability of the algorithm to infer the hidden states assignment:
\begin{equation}
    ACC = \frac{\sum_i\mathbbm{I}(\hat{d_i} = d_i)}{N}=1-P_e
\end{equation}
where $\hat{d}_i$ is the estimated index of the $i$th observation, and $d_i$ is the true index of the $i$th observation. The Munkres algorithm \cite{munkres1957algorithms} is used to map randomly selected indices of the estimated index sequence to the set of indices that maximize the overlap with the true index sequence.

\textbf{MSE}: The MSE indicates the ability to estimate the parameter value of the HMPs, in this paper, we estimate the mean of symbols:
\begin{equation}
    MSE = \frac{1}{MK}\sum_{m=1}^M \sum_{k=1}^K \mathbb{E}[(\hat{\boldsymbol{\mu}}^m_k-\boldsymbol{\mu}^m_k)^2]
\end{equation}
where $\hat{\boldsymbol{\mu}}^m_k$ is the estimated parameter value and $\boldsymbol{\mu}^m_k$ is the true parameter value.

\subsection{Performance Validation}
De-interleaving accuracy and parameter estimation accuracy are evaluated in this part. Metrics are computed on a per-dataset basis and averaged over 100 Monte Carlo simulations. Algorithms were run for a maximum of 100 iterations or until convergence. The parameters $\boldsymbol{\Gamma}$ are randomly initialized.


\subsubsection{Comparisons}
Three de-interleaving methods are used as baseline methods in this study:

\begin{enumerate}
  \item [1.] \textbf{Genetic Algorithm} (GA) is a heuristic optimization method given in \cite{Holland}. The objective is set as the penalized maximum likelihood defined in \eqref{penalized ML}.
  \item [2.] \textbf{Gaussian Mixture Model} (GMM) is an unsupervised clustering algorithm given in \cite{Jordan1994Hierarchical}. The GMM-based method has been applied in the radar signal de-interleaving \cite{Scherreik2021Online}.
  \item [3.] \textbf{Hidden Markov Model} (HMM) is described in \cite{mor2021systematic}. The HMM can be treated as a GMM with time dependencies \cite{Kontorovich2013}. The hidden state assignments are the de-interleaving results. 
\end{enumerate}

\subsubsection{De-interleaving Accuracy}

\begin{figure}[!t]
  \centering
  \includegraphics[width=3.6in]{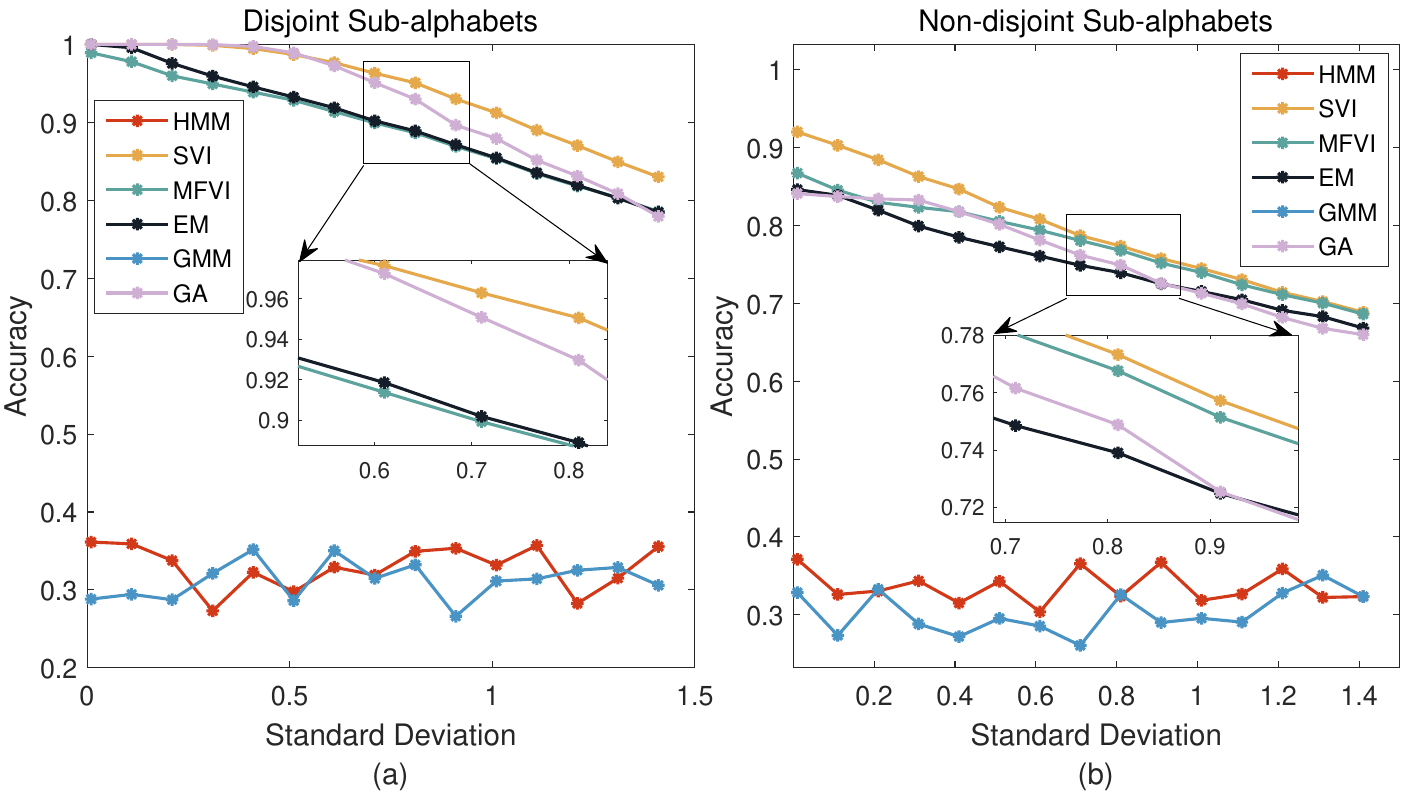}
  \caption{De-interleaving results of various de-interleaving algorithms.}
  \label{de-interleaving result}
\end{figure}

In this part, we test the de-interleaving accuracy under various SDs. The SD $\sigma$ is increased from 0 to 1.5 with a step of 0.1. The de-interleaving accuracy results for disjoint sub-alphabets are shown in Fig.~\ref{de-interleaving result}(a). The baseline methods (GMM and HMM) show poor performance due to the model mismatch, and the de-interleaving accuracy is under 0.4. Among all the tested algorithms, SVI has the best performance under all SD values. The performance of these algorithms (EM, MFVI, GA, SVI) decreases as SD grows. The de-interleaving accuracy for non-disjoint sub-alphabets is shown in Fig.~\ref{de-interleaving result}(b), when the sub-alphabets are non-disjoint, the de-interleaving accuracy is slightly lower than the scenario with disjoint sub-alphabets, but it still has about 90\% de-interleaving accuracy when SD close to zero. 

Two phenomena need to be explained. Firstly, though GA has outstanding performance, GA consumes a lot of time for optimization. The time-consuming nature makes it unpractical. Secondly, the exact EM algorithm does not present the best performance we anticipated. The reason for the unsatisfactory performance of the EM algorithm is that its state space is large, leading to higher possibilities for local optima compared with other algorithms. The variational inference method can effectively reduce the size of the state space, thereby speeding up convergence. 

\subsubsection{Parameter Estimation Accuracy}
\begin{figure}[!t]
  \centering
  \includegraphics[width=3.6in]{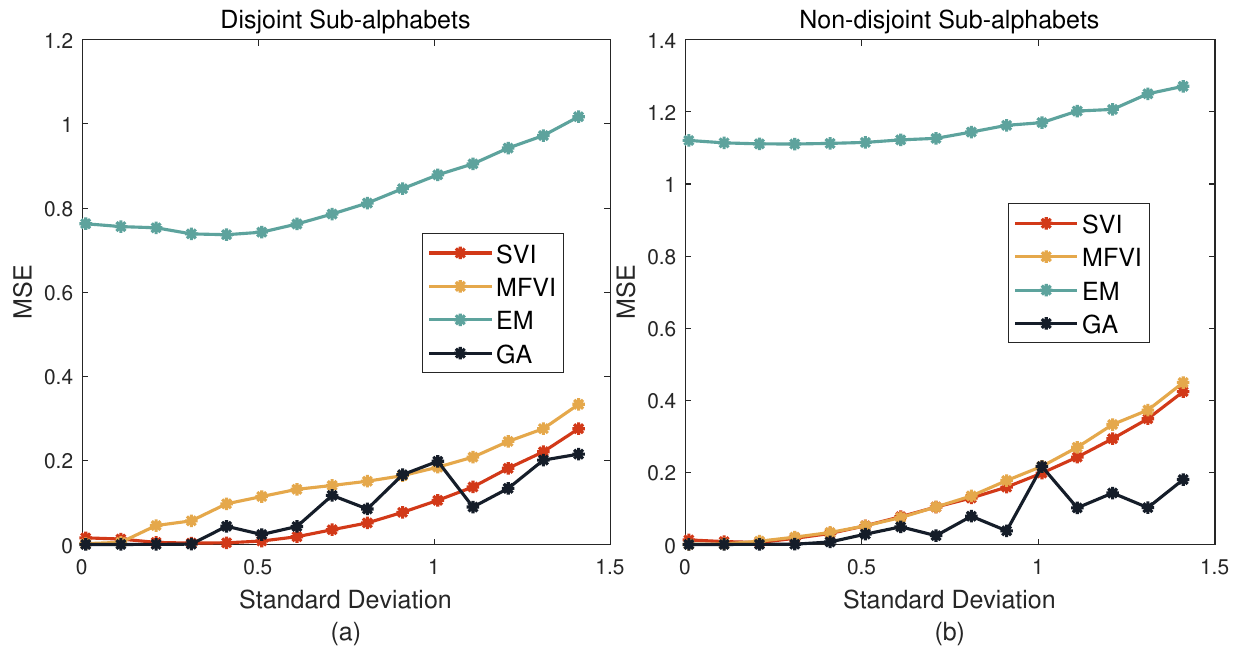}
  \caption{Parameter estimation results of various de-interleaving algorithms.}
  \label{MSE result}
\end{figure}

It is concluded from the previous part that GMM and HMM can not de-interleave the mixtures of HMPs accurately. To further demonstrate methods with good performance, the GMM and HMM results are omitted in this part. We test the parameter estimation accuracy by varying the SD values. The SD $\sigma$ is increased from 0 to 1.5 with a step of 0.1, results shown in Fig.~\ref{MSE result}. It can be seen that both algorithms can have effective results. Specifically, the approximate methods are more accurate than the exact EM method, the reason is that the approximate methods drop some couplings of the graphical model to restrict the size of state space to achieve better performance. Though the EM has the highest MSE among these four algorithms, the MSE value is around 0.8 facing the disjoint sub-alphabets. As for the non-disjoint sub-alphabets, the MSE value is lower than 1.4. In summary, the result shows the effectiveness of the proposed algorithms. Among the four algorithms, GA presents the superior performance and the estimation result is more accurate than the proposed algorithms under high standard deviations.

\subsection{Performance under missing observations}
\begin{figure}
  \centering
  \includegraphics[width=3.4in]{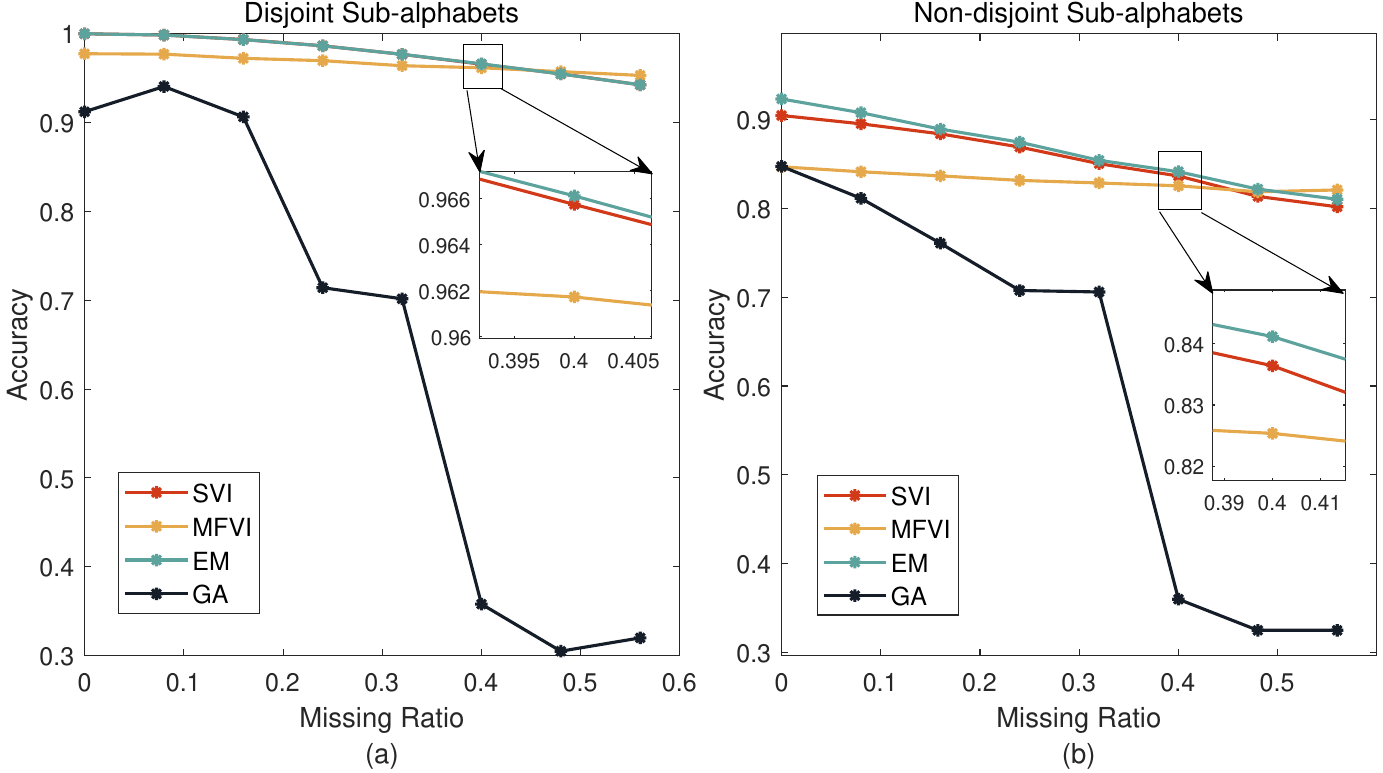}
  \caption{De-interleaving results of the de-interleaving algorithms under random missing observations.}
  \label{missing result}
\end{figure}

In this part, the robustness under random missing observations is tested. The standard deviation is fixed as 0.1 and the missing ratio increased from 0 to 56\% with the step of 0.8, resulting in 8 different missing situations. Simulation results are shown in Fig.~7. The EM and SVI have similar performance under missing observations, and the de-interleaving accuracy of the three proposed algorithms is close to 100\% as shown in Fig.~\ref{missing result}(a). The proposed algorithms are equipped with superior performance because we model the switching process by a one-order hidden Markov chain. Missing observations would result in slightly changing the state transition matrix, the de-interleaving accuracy would not change unless there are too many missing observations to change the transition tendency. The state transition tendency would not change unless the proportion of the observation data changes significantly. The simulation results in Fig.~\ref{missing result}(b) verified that the performance of the proposed methods under non-disjoint sub-alphabets is effective. 

In either disjoint or non-disjoint sub-alphabets situations, the GA has huge performance degradation when the missing ratio is higher than 0.2. Specifically, for the disjoint sub-alphabets scenario, the performance dropped from 0.9 to 0.3 for the missing ratio increased from 0.2 to 0.56; in terms of the non-disjoint scenario, the performance decreased from 0.85 to 0.3 with the increase of the missing ratio. This comparison verified that the GA is not robust to the missing observation situations.

\section{Applications}
In this section, we apply our algorithms to two applications including de-interleaving mixtures of radar pulse sequences and human motion separation.
\subsection{De-interleaving mixtures of Radar Pulse Sequences}

De-interleaving mixtures of radar pulse sequences is typically performed on Pulse Descriptor Words (PDWs) comprising the Time of Arrival (TOA), Angle of Arrival (AOA), Radio Frequency (RF), Pulse Width (PW), and Pulse Repetitive Interval (PRI). Major de-interleaving techniques rely on the clustering of PDWs based on subsets of these parameters \cite{WangChao2022,Scherreik2021Online}. However, The AOA and TOA estimation would suffer performance degradation under low SNR \cite{Zhang2023}, while other PDWs like RF and PW may have large SD. The de-interleaving scheme may only be performed solely based on the PDWs with high SD and pulse arriving orders. Meanwhile, with the development of agile radar \cite{Huang2018analysis}, a radar emitter may vary its transmissions and exhibit multiple symbols, which brings a great challenge for de-interleaving and recognizing the radar work mode. Existing methods only use either timing information (PRI, TOA) or inter-pulse modulations (PW, RF, AOA) to perform radar signal de-interleaving. The proposed methods combine the inter-pulse modulation and timing information to achieve better performance. Without the loss of generality, we de-interleave mixtures of radar pulse sequences using RF and pulse arrival order, but the proposed methods can be extended to other PDWs and multi-parameters case.

When multiple radars use different RF values, the problem is reduced to de-interleave when sub-alphabets are disjoint, the effectiveness and robustness under missing observations are verified in the previous section. In this part, we consider a relatively extreme situation, the model identical radar. Model identical radar means multiple radars have the same PDWs. The only difference between the two model identical radars is that the initial phases are different. The initial phase represents the time of the first radar pulse that was intercepted. We examined the influence of the initial phase.

Two model identical radars utilized two RF values, and the observation of each value follows a Gaussian distribution due to the observation noise, the symbols are denoted by $\boldsymbol{\Theta} =\{(\mu_1,\sigma_1^2),(\mu_2,\sigma_2^2)\}$. Specifically, we set $\boldsymbol{\Theta} = \{(1245, 1^2),(1230, 1^2)\}$, where the values are in $kHz$ and $kHz^2$. The emission time of each radar uses jittered modulation\footnote{The emission time uses jittered modulation, which means that each radar has pulse repetitive interval (PRI) follows Gaussian distribution, details are described in \cite{Bao2023}.}. The mean and variance of the jittered PRI is 50$\mu s$ and 0.8$\mu s^2$. The initial phase of the first radar is uniformly distributed over $[0, 10]\mu s$. The initial phase of the second radar is gradually farther from the first radar (i.e., uniformly distributed over $[\alpha, \alpha +10] \mu s$, wherein $\alpha\in[0,20]$ is the overlapping level). Varying the initial phase, the de-interleaving scenarios of model identical radar are obtained. Three approaches were implemented to de-interleave signals emitted by two model identical radars:
\begin{enumerate}
  \item Structured Variational Inference (SVI): the SVI is implemented by optimizing \eqref{structureLowerBound}.
  \item Expectation Maximization (EM): the EM is implemented by optimizing \eqref{Qfunction}.
  \item Neural Translation Network (NMT): the NMT de-interleaving framework was first proposed in \cite{Zhu2022}. The parameters of the network were optimized in a supervised manner.
\end{enumerate} 


The results of the de-interleaving model identical radar are shown in Fig. 8(a). With the increase of $\alpha$ value, the de-interleaving accuracy increases. It can be seen from the result that the accuracy of SVI and EM achieve 0.9 as $3
\leq \alpha \leq 20$, and SVI increases faster than EM. The accuracy of the NMT achieves 0.9 as $13 \leq \alpha \leq 20$. The proposed algorithms have superior performance when the two initial phases are close to each other. When the initial phases are far away from each other, the NMT has superior performance, thanks to the supervised manner. 

\begin{figure}
  \centering
  \includegraphics[width=3.56in]{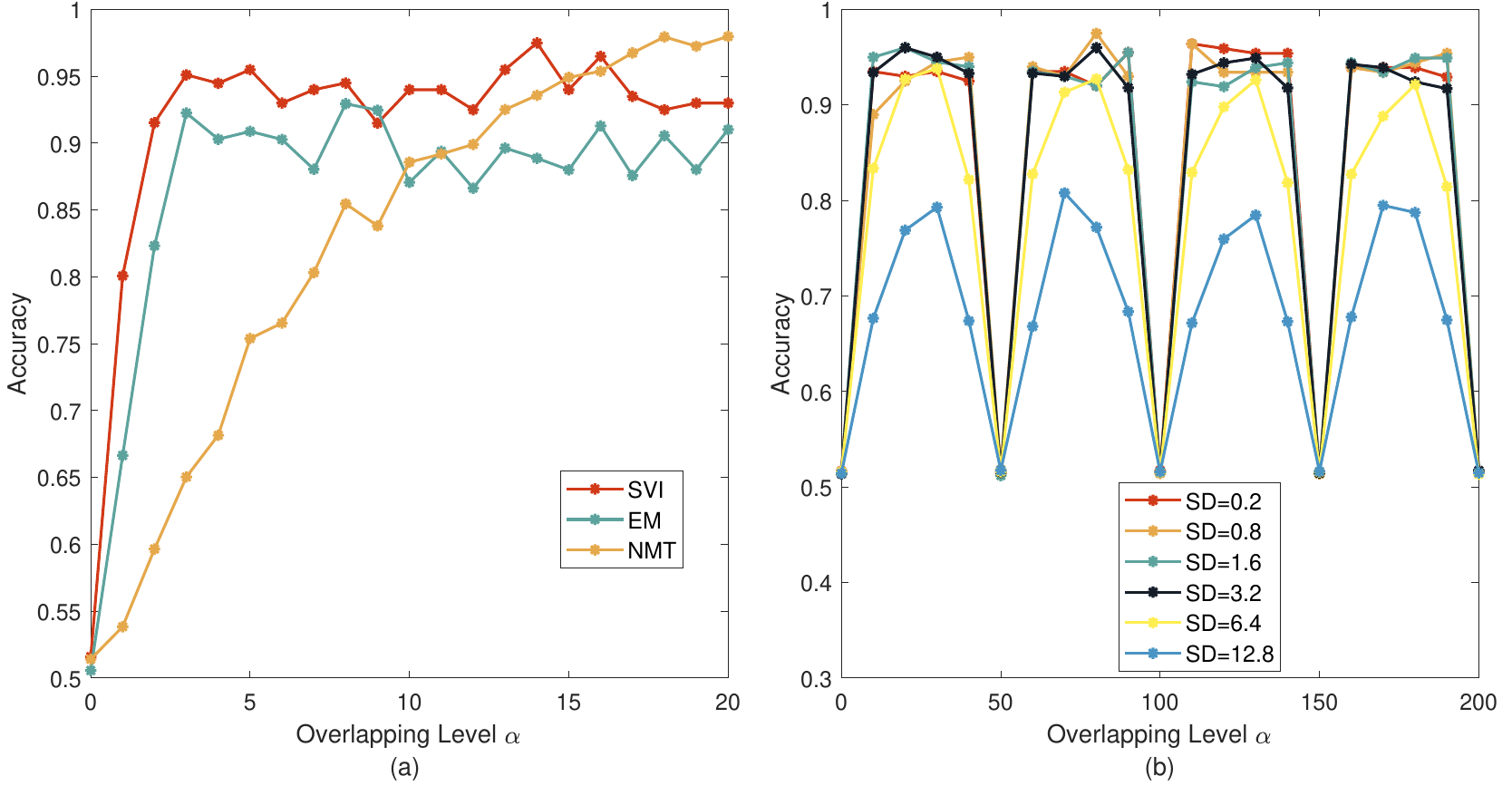}
  \caption{De-interleaving results of model identical radar data. (a)The comparisons between SVI, EM and NMT when $0\leq \alpha\leq 20$. (b)The comparison of various jittered deviations when $0\leq \alpha \leq 200$ using SVI.}
  \label{radar data result}
\end{figure}

It can be concluded that the SVI performs better than EM. We then verify the robustness of the proposed SVI method to different jittered PRI deviations. We set the jittered PRI deviation gradually increased from 0.2$\mu s^2$ to 12.8$\mu s^2$ and the overlapping level $\alpha$ is increased from 0 to 200. The results are shown in Fig.~\ref{radar data result}(b). From the perspective of different jittered PRI deviations, the de-interleaving accuracy decreases when the jitter deviation increases. It is worth mentioning that the SVI method has a maximum of 0.8 accuracy when the jitter deviation is 12.8$\mu s^2$. From the perspective of overlapping level $\alpha$, the performance shows a periodic trend among values of multiples of 50 (i.e., $\alpha$ = 50, 100, 150, 200) since the mean of the jitter value is 50$\mu s$. The phenomenon can be explained intuitively. When the initial phase of the second radar is infinitely close to that of the first radar, since we only use the order of arrival as available information, the algorithm will completely be confused, resulting in a de-interleaving accuracy close to 0.5. That means the pulses are randomly de-interleaved.

\subsection{Human Motion Separation}
In the application of distributed systems, there is an architecture of distributed sensing and centralized processing. In practice, we can not always ensure that the data is transmitted to the data center synchronously. Asynchronous transmission means that the received data is anonymized or unlabeled. The data center needs to label the received data. In this part, we assume that two sensors monitor the movements of two people. The data is obtained from the dataset of HASC challenge 2011 dataset \cite{kawaguchi2011hasc2011corpus}. The three-dimensional data was converted to one-dimensional data via l2 norm, similar transformation is also performed in \cite{de2021change}. 

\begin{figure}
  \centering
  \includegraphics[width=3.56in]{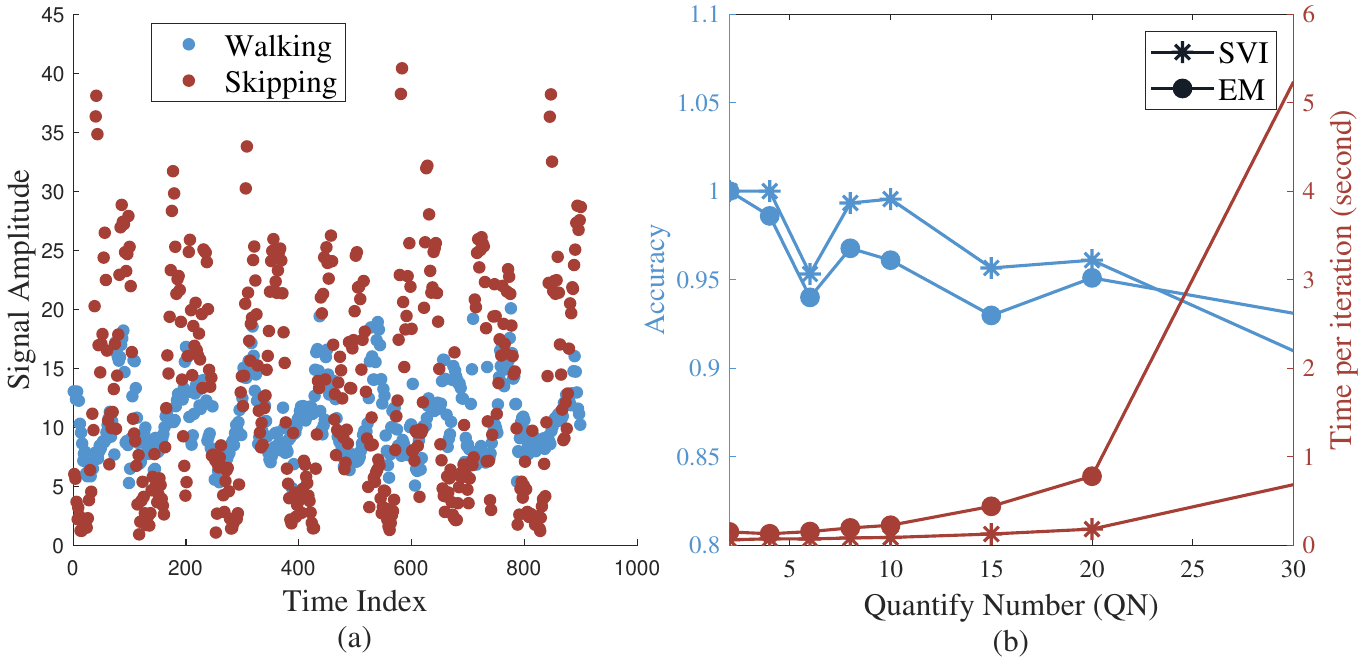}
  \caption{De-interleaving results of human motion data. (a) The de-interleaving results when QN=2 on unlabeled human motion data via SVI. (b) The de-interleaving accuracy of SVI and EM algorithm and their relative time per iteration on an Intel Core i9-10850K CPU running Matlab.}
  \label{Human motion data}
\end{figure}

In our simulation, one person is skipping while the other is walking, their motions are collected by two sensors with the same sampling rate, quantified, and randomly interleaved at the data center. We hereby define Quantify Number (QN), the QN refers to the number of possible values obtained by the sensor after quantizing the data. In this paper, the QN corresponds to the hidden state number of the component chain. The interleaved data separated by SVI and EM are described in this paper. The de-interleaving results when QN=2 are shown in Fig.~\ref{Human motion data}(a). The activities of two people are successfully separated.

To find out the influence of the QN setting, we increase the QN value from 0 to 30. The de-interleaving accuracy and the running time per iteration are recorded and depicted in Fig.~\ref{Human motion data}(b). From the perspective of the de-interleaving accuracy, with the increase of the QN, the accuracy is slightly decreased as the QN increase. When the QN is large, the quantified value of two sensors may overlap, resulting in non-disjoint sub-alphabets causing performance degradation. In terms of the processing time per iteration, when the QN increases, it can be verified that the relative time per iteration of the EM method rises faster than the SVI method. The detailed complexity analysis is discussed in Section IV.

\section{Conclusion}
In this paper, we have studied the problem of representation and de-interleaving of the mixtures of the hidden Markov processes. Firstly, we designed a generative model to represent the IHMP. Secondly, we proposed an exact algorithm (EM) and two approximate algorithms (MFVI and SVI) to infer the hidden variables of the generative model. Finally, an error probability lower bound was derived based on the likelihood ratio test.

The simulations demonstrate that the SVI method performs better than other proposed methods and baseline methods under both disjoint and non-disjoint sub-alphabets. The effectiveness of the proposed methods was verified through radar data and human motion data. Besides, there is still room for further developments. In this paper, the inference of the generative model is performed requiring the hidden state numbers $M, K$, and the Bayesian non-parametric prior can be applied to infer the hidden state numbers automatically.

\section*{Appendix}
\subsection{The M-step}
Taking the derivative of the \eqref{Qfunction} with respect to $\boldsymbol{\mu}^m$, and set it equal to zero:
\begin{equation}
  \begin{aligned}
    &\frac{\partial \mathbb{Q}}{\partial \boldsymbol{\mu}^m} = \\
    &\sum_{t=1}^T \bigg\{\sum_{n=1}^M {\boldsymbol{\mu}^n} \mathbb{E}(Z_{t,n}{\boldsymbol{S}_t^n}^\top Z_{t,m}\boldsymbol{S}_t^m) - \mathbb{E}(Z_{t,m}\boldsymbol{S}_t^m)\boldsymbol{p}_t\bigg\}
  \end{aligned}
\end{equation}
Solving the above equation, and $\boldsymbol{\mu}^m$ is updated by the following equation:
\begin{equation}
  \boldsymbol{\mu}^m = \bigg(\sum_{t=1}^T \mathbb{E}(\boldsymbol{S}_{t,m}) \boldsymbol{p}_t\bigg)\bigg(\sum_{t=1}^T \mathbb{E}(Z_{t,n}{\boldsymbol{S}_t^n}^\top Z_{t,m}\boldsymbol{S}_t^m)\bigg)^\dagger
\end{equation}
where $\dagger$ is the Moore-Penrose pseudo-inverse. The update function of $\boldsymbol{\Sigma}$ is obtained by similar progress, the derivation is omitted in this paper. The calculation of the $\boldsymbol{\pi},\boldsymbol{\pi}^z,\boldsymbol{A},\boldsymbol{A}^z$, and E-steps are the standard forward-backward algorithm of a Markov chain, detail refers to \cite{Bishop2006}.

\subsection{The Derivation of MFVI}
Using the lower bound defined in \eqref{lowerbound}, we take the expectation of the logarithm of the joint probability of IHMP \eqref{JointProbability} is rewritten  as:
\begin{equation}
  \label{logJoint}
  \begin{aligned}
    &\mathbb{E}_Q(\log P(\boldsymbol{Z},\boldsymbol{S},\boldsymbol{\Gamma},\boldsymbol{p})) \propto -\frac{1}{2}\sum_{t=1}^{T} \bigg\{\boldsymbol{p}_t^\top\boldsymbol{\Sigma}^{-1}\boldsymbol{p}_t \\
    &-2\boldsymbol{p}_t\boldsymbol{\Sigma}^{-1}\left(\sum_{m=1}^{M}\phi_{t,m}{\boldsymbol{\theta}_t^m}^\top\mu^m\right)^\top\\
    &+\sum_{m=1}^M\sum_{n=1}^M tr\{\boldsymbol{\Delta}_m^n(\phi_{t,m}\boldsymbol{\theta}_t^m)(\phi_{t,n}\boldsymbol{\theta}_t^n)^\top\}
    \bigg\}\\
    &+\sum_{m=1}^{M} {\boldsymbol{\theta}_1^m}^\top \log \boldsymbol{\pi}^m+\sum_{t=2}^T \sum_{m=1}^{M}{ \boldsymbol{\theta}_t^m }^\top \mathbb{E}_Q(\bar{\boldsymbol{A}}^m_t) \boldsymbol{\theta}_{t-1}^m\\
    &+\boldsymbol{\phi}_1^\top \log \boldsymbol{\pi}^z 
    +\sum_{t=2}^{T} \boldsymbol{\phi}_t^\top \log \boldsymbol{A}^z \boldsymbol{\phi}_{t-1}
  \end{aligned} 
\end{equation}
The variational distribution \eqref{MF} is written as:
\begin{equation}
\begin{aligned}
&\mathbb{E}_Q(\log Q(\boldsymbol{S},\boldsymbol{Z},\boldsymbol{\Gamma})) = \\
  &\sum_{t=1}^{T}\sum_{m=1}^{M} {\boldsymbol{\theta}_t^m}^\top \log \boldsymbol{\theta}_t^m +\sum_{t=1}^{T}{\boldsymbol{\phi}_t}^\top \log \boldsymbol{\phi}_t
\end{aligned}
\end{equation}

The lower bound $\mathcal{L}=\mathbb{E}_Q(\log P(\boldsymbol{Z},\boldsymbol{S},\boldsymbol{\Gamma},\boldsymbol{p}))-\mathbb{E}_Q(\log P(\boldsymbol{S},\boldsymbol{Z},\boldsymbol{\Gamma}))$ is formulated as \eqref{MF_ELBO}. Taking derivatives with respect to $\boldsymbol{\theta}_t^m$ and $\phi_{t,m}$, then set it equal to zero, the update function \eqref{mfupdatetheta} and \eqref{mfupdatephi} is obtained.

\subsection{The Derivation of SVI}
Derivation of the SVI method follows a similar step as MFVI described in the above part. The expectation of the logarithm of the joint probability of the generative model is the same as \eqref{logJoint}. The logarithm expectation of the structured variational probability can be written as:
\begin{equation}
  \begin{aligned}
    &\mathbb{E}_Q(\log Q(\boldsymbol{S,Z,\Gamma})) \propto\\
    & +\sum_{m=1}^{M} {\boldsymbol{\theta}_1^m}^\top \log \boldsymbol{\pi}^m + {\boldsymbol{\phi}_1}^\top \log \boldsymbol{\pi}^z \\
    &+ \sum_{t=2}^{T} \sum_{m=1}^{M} (\boldsymbol{\theta}_t^m \log \boldsymbol{A}^m \boldsymbol{\theta}_{t-1}^m) 
    + \sum_{t=1}^{T} \sum_{m=1}^{M} \boldsymbol{\theta}_t^m \log \boldsymbol{h}_t^m\\
    &+\sum_{t=2}^{T} {\boldsymbol{\phi}_t}^\top \log \boldsymbol{A}^z \boldsymbol{\phi}_{t-1}+ \sum_{t=1}^{T} \boldsymbol{\phi}_t^\top \log \boldsymbol{g}_t
  \end{aligned}
\end{equation}

The lower bound $\mathcal{L}=\mathbb{E}_Q(\log P(\boldsymbol{Z},\boldsymbol{S},\boldsymbol{\Gamma},\boldsymbol{p}))-\mathbb{E}_Q(\log P(\boldsymbol{S},\boldsymbol{Z},\boldsymbol{\Gamma}))$ is then formulated as \eqref{structureLowerBound}. Taking derivatives with respect to $\log \boldsymbol{h}_t^m$ and $\log g_{t,m}$, and setting it equal to zero, the update functions \eqref{structureupdatetheta} and \eqref{structureupdatephi} were obtained.

\subsection{The Proof of Theorem 2}
When $x=2$, $y=1$, we first calculate the probability of error event $\mathcal{E}_i^{21}$. According to the stationary assumption, we omit the initial phase of each Markov chain in \eqref{JointProbability}, the stationary Markov chain with Gaussian distribution can be seen as a mixture of distributions. The probability density function of the $m$th HMM can be rewritten as:
\begin{equation}
    P(\boldsymbol{p}|\boldsymbol{\varphi}_i^m)\sim\sum_{k=1}^K \xi^m_k f_{\boldsymbol{\varphi}_i^m}(\boldsymbol{p})
\end{equation}
where $\xi_k^m$ is the stationary distribution as well as the mixture coefficient. Thus, we can rewrite the log-likelihood of the generative model of the IHMP:
\begin{equation}
    \begin{aligned}
        f^m(p) =&\log {A}^z_{2,m} +{\Xi^m}^\top \log \boldsymbol{A}^m \boldsymbol{c}^m+  \sum_{k=1}^2 \log \xi_k^m\\
        &+\sum_{k=1}^2 \left\{ c_k^m \left[\log \frac{1}{\sqrt{2\sigma^2}}-\frac{1}{2\sigma^2}(p-\mu_k^m)^2 \right]\right\}  
    \end{aligned}
\end{equation}
where $\log\xi_k^m$ arises from the mixture coefficient and $\boldsymbol{c}^m = \{c_k^m\}_{k=1}^2$ is the indicator variable of the $m$th component chain. In our derivation, we do not know the hidden state of the source $x$ at time instant $i-1$, thus the transition term is reduced to ${\Xi^m}^\top \log A^m \boldsymbol{c}^m$. An error event $\mathcal{E}^{12}_i$ happens if $f^2(p)>f^1(p)$. Equivalently, 
\begin{equation}
\begin{aligned}
        &2p\sum_{k=1}^2(c_k^2\mu_k^2-c_k^1\mu_k^1)>\\
        &\sum_{k=1}^2 2\sigma^2\log\frac{\xi_k^1}{\xi_k^2}+2\sigma^2\log\frac{{A}^z_{2,1}}{{A}^z_{2,2}} + \sum_{k=1}^2({c_k^2\mu_k^2}^2-{c_k^1\mu_k^1}^2)\\
        &+2\sigma^2{\Xi^1}^\top \log \boldsymbol{A}^1 \boldsymbol{c}^1-2\sigma^2{\Xi^2}^\top \log \boldsymbol{A}^2 \boldsymbol{c}^2
\end{aligned}
\end{equation}

There are $4!=24$ scenarios, apart from the scenarios of symmetry and restrictions according to the assumption described in theorem 2 (i.e., $\mu_1^i<\mu_2^i$ and $\mu_1^1<\mu_2^2$), there are $\frac{4!}{2!\prod_{i}2!}=3, i\in[1,2]$, i.e., $\mu_1^1>\mu_2^1>\mu_1^2>\mu_2^2$, $\mu_1^1>\mu_1^2>\mu_2^1>\mu_2^2$ and $\mu_1^1>\mu_1^2>\mu_2^2>\mu_2^1$. In each scenario, there are total of four cases of $c_k^m$ can be chosen to produce the error event $\mathcal{E}_i^{21}$: $[c^1_1, c^2_1; c^2_1,c^2_2]\in\{[1,0;1,0],[1,0;0,1],[0,1;1,0],[0,1;0,1]\}$. In the first scenario, the error probability lower bound is formulated as:
\begin{equation}
  P(\mathcal{E}_i^{21}) = \sum_{k=1}^{2}\sum_{l=1}^{2} \xi_k^1\xi_l^2P(p>\gamma_{kl}^{21})
\end{equation}
where,
\begin{equation}
\begin{aligned}
     &\gamma_{kl}^{21} = \frac{({\mu_k^2}^2-{\mu_l^1}^2)}{2(\mu_k^2-\mu^1_l)}+\\
     &2\sigma^2\frac{\log\frac{\xi_k^2{A}^z_{2,1}}{\xi_k^1{A}^z_{2,2}} + ({\Xi^1}^\top \log \boldsymbol{A}^1_{\cdot,k}-{\Xi^2}^\top \log \boldsymbol{A}^2_{\cdot,l})}{2(\mu_k^2-\mu^1_l)}
\end{aligned}
\end{equation}

The observation variable is generated from a Gaussian mixture model, and the probability $P(p>\gamma_{kl}^{21})$ is calculated by the right tail function. Specifically,
\begin{equation}
  \begin{aligned}
    &P(p>\gamma_{kl}^{21}) = \xi_k^1\mathcal{Q}\bigg(\frac{{\mu}_k^1-\gamma_{kl}^{21}}{\sigma}\bigg)\\
  \end{aligned}
\end{equation}

Similarly, we can explicitly express the probability of the other error events $P(\mathcal{E}_i^{xy})$, as well as other scenarios. By summarizing these scenarios, we finally obtain a unified expression illustrated in \eqref{lower bound}.
\label{sec:conclusion}

\bibliographystyle{IEEEtran}
\bibliography{ref}




\end{document}